\documentclass[11pt,twoside]{article}
\title{On a non-local spectrogram for denoising one-dimensional signals
\thanks{Supported by the Spanish MCINN Project
MTM2010-18427. }}

\author{Gonzalo Galiano  \thanks{Dpt. of Mathematics, Universidad de Oviedo,
 c/ Calvo Sotelo, 33007-Oviedo, Spain ({\tt galiano@uniovi.es, julian@uniovi.es})}
    \and Juli\'an Velasco\footnotemark[2] }
\date{}
\usepackage{fancyhdr}
\usepackage{amsmath}
\usepackage{amssymb}
\usepackage{graphicx}
\usepackage{multirow}

\newcommand{\drop}[1]{}
\newcommand{\no}{\noindent}
\newcommand{\fer}[1]{(\ref{#1})}
\newcommand{\qtext}[1]{\quad\text{#1}}

\newcommand{\bx}{\mathbf{x}}
\newcommand{\by}{\mathbf{y}}
\newcommand{\bz}{\mathbf{z}}

\newcommand{\eps}{\varepsilon}
\newcommand{\vfi}{\varphi}

\newcommand{\grad}{\nabla}
\newcommand{\om}{\omega}

\newcommand{\R}{\mathbb{R}}

\def\O{\Omega}

\newcommand{\G}{{\mathcal{G}}}

\newcommand{\abs}[1]{| #1 |}
\newcommand{\nor}[1]{\| #1 \|}

\DeclareMathOperator{\mean}{Mean} 
\DeclareMathOperator{\INT}{INT} \DeclareMathOperator{\IF}{IF}
\DeclareMathOperator{\SNR}{SNR} \DeclareMathOperator{\Div}{div}
\DeclareMathOperator{\WV}{WV}

\begin{document}

\maketitle

\begin{abstract}
In previous works, 
we investigated the use of 
local filters based on partial differential 
equations (PDE) to denoise one-dimensional signals through the image processing of 
time-frequency representations, such as the spectrogram. 
In this image denoising algorithms, the particularity 
of the image was hardly taken into account. We turn, in this paper, to study the performance 
of non-local filters, like Neighborhood or Yaroslavsky filters, in the same problem. We show
that, for certain iterative schemes involving the Neighborhood filter, the computational time is 
drastically reduced with respect to Yaroslavsky or nonlinear PDE based filters, while the outputs
of the filtering processes are similar. This is heuristically justified by the connection between
the (fast) Neighborhood filter applied to a spectrogram and the corresponding Nonlocal Means filter 
(accurate) applied to the Wigner-Ville distribution of the signal. This correspondence holds only 
for time-frequency representations of one-dimensional signals, not to usual images, and in this sense 
the particularity of the image is exploited. We compare though a series of experiments on synthetic and
biomedical signals the performance of local and non-local filters.
\end{abstract}

\begin{itemize}
 \item[] {\small \emph{Keywords: }
Spectrogram, image restoration, local and non-local filtering, instantaneous frequency, biomedical signals.
} 

\end{itemize}

\section{Introduction}
Denoising one-dimensional signals is an important topic which is usually addressed from filter theory in time or frequency domains. In some applications in which processing speed is not a fundamental issue, filters defined in the joint time-frequency domain may be considered, usually improving the filtering process. Examples of this situation are found in Electrocardiogram (ECG) and other biomedical signals \cite{sornmo2005}, human voice analysis \cite{xu2008} or animal sound analysis \cite{hopp1998}. 

With respect to the latter, in previous works \cite{dfg2007,dfgv2007b,dfgv2008} we investigated the use of time-frequency distributions to estimate the number of wolves howling in a giving recording to provide an estimation of the number of individuals in a pack. This estimation is the basis for counting regional wolf populations which is of interest for both ecological and economic purposes, since  authorities must reimburse the cost of cattle killed by this protected specie \cite{skonhoft}. Of course, and despite the quality of recording devices, field recordings are usually affected for a variety of undesirable signals which range from low amplitude broad spectrum long duration signals, like wind, to
 signals localized in time, like cattle bells, or localized in
spectrum, like car engines. Clearly, the addition of all these
signals generates an unstructured noise in the background of the
wolves chorus which must be treated for a proper signal analysis.

Medical signals are another good example of this situation. Due to the electromagnetic fields created by measuring devices, the usual low frequency signals to be acquired are contaminated by a background noise which is usually in the same frequency band that the signal of interest. Therefore, fine denoising techniques must be applied to segregate the signal of interest form the noise.

In general, the denoising procedure is not aimed to recovering a clean signal but to produce a clean time-frequency representation of the signal which allows further analysis techniques, for instance and importantly, the instantaneous frequency (IF) estimation.
For the examples given above, IF estimation allows to count the number of different individuals howling in the recording (each individual being identified with an IF line). We also provide an example in which the spectrogram energy content of 
an ECG signal is filtered to identify an arrhythmia episode.

In \cite{dfg2007,dfgv2007b,dfgv2008b}, we used nonlinear diffusion image denoising techniques applied to the spectrogram of a sound signal, a wolf chorus. Although, as above mentioned, execution time is not a relevant issue for this type of problems, we found that nonlinear diffusion algorithms require a high computational time, making their use not operative in many situations. In addition, these filters do not take advantage of the special characteristics of the image produced from the spectrogram, i.e. they operate on the spectrogram as in any other image. In this article 
we show that nonlocal filters such as Neighborhood filters \cite{lee1983} are computationally more efficient to deal with these images and give similar results. Moreover, we point out a relationship between the Nonlocal Means filter \cite{buades2005,buades2010} and the Neighborhood filter which is exclusive of their implementation on images defined through time-frequency distributions.

The outline of the article is as follows. We present in Section 2 the mathematical framework of the problem and the filtering techniques proposed in this article for one-dimensional signal denoising. In particular, we justify our choice of the Neighborhood filter as an inexpensive approximation to the well known Nonlocal Means filter for the special case of spectrogram images. In Section 3, we introduce the discrete problem and deduce the corresponding formulas for algorithm 
implementation. Apart from the Neighborhood filter, we consider the Yaroslavsky-SUSAN
\cite{yarolavsky1985,smith1997} filter and a nonlinear diffusion filter based in the Total Variation norm \cite{rudin1992,dfgv2008b}, for comparison porpouses. Then, we demonstrate the performance of these filters by applying them to three noisy signals (synthetic, wolf chorus and ECG) and give quantitative comparisons based on the Mean Square Error (MSE), and the visual inspection of the processed spectrograms and other related magnitudes.

\section{Mathematical framework}

Let $f\in L^2(\R)$ denote a one-dimensional signal and 
$\WV(f;\cdot,\cdot)$ be its Wigner-Ville distribution, defined as 
\[
\WV ( f ; t , \om ) = \int_\R f( t + \frac{s}{2} )  \bar{f}(t -
\frac{s}{2})e^{-iws} ds,
\]
where $\bar{f}$ denotes the complex conjugate of $f$. 
The Wigner-Ville distribution has received much attention for
IF estimation due to its excellent concentration for mono-signals and many other
desirable mathematical properties, see \cite{mallat}. However, it
is well known that it presents high amplitude sign-varying
cross-terms for multi-component signals which makes its
interpretation difficult. 
For attenuating these interference terms several approaches have been followed, 
mainly based on the smoothing of the $\WV$ by convolution  with a suitable regularizing kernel.
Special mention is due to the spectrogram, which may be defined either as the 
energy density function of the short time Fourier transform
\begin{equation}
\label{gabor}
 \G_\vfi (f;t,\om)=\int_\R f(s)\vfi(s-t)e^{-i\om s}ds,
\end{equation}
for some real, symmetric and normalized 
 window $\vfi\in L^2(\R)$, i.e.
\begin{equation}
\label{spec}
 S_\vfi(f;t,\om)= \G_\vfi (f;t,\om) \bar{\G}_\vfi (f;t,\om) ,
\end{equation}
or as the convolution product of the Wigner-Ville distributions of the signal and the window
\begin{equation}
\label{specwv}
S_\vfi(f;t,\om)=\int_{\R^2} \WV(\vfi;\tilde t,\tilde \om) \WV(f;t-
\tilde t,\om- \tilde \om) d\tilde t d \tilde\om.
\end{equation}
In practice, for finite time duration and band-limited one-dimensional signals the time-frequency domain is bounded. Let us denote this domain by $\O$, and write a generic time-frequency point $(t,\om)\in \O$ by $\bx$. Dropping the dependence on the signal $f$ and the window $\vfi$ in the notation for the spectrogram, the conservation of energy implies $\nor{S}_{L^2(\O)}=\nor{f}_{L^2(0,T)}$, and the 
additional assumption $f,\vfi\in L^\infty(0,T)$ implies  $S\in L^\infty(\O)$ as may be easily deduced from \fer{gabor}-\fer{spec}. Therefore, after a suitable normalization, we may consider the spectrogram of a signal as an image and apply to it well-known techniques of image denoising and enhancing for the estimation of the underlaying signal. In addition, the definition of $S$ as a convolution product involving the window $\vfi$ ensures further regularity when $\vfi$ is smooth. For instance, for Gaussian windows, $S\in C^\infty (\O)$.

First attempts to denoising and enhancing of spectrogram images were made via local differential filters \cite{dfg2007,dfgv2007b,dfgv2008b}
based on corresponding well established methods, see \cite{alm1992,auger1997,rudin1992}, in which the filtering process of a intensity image $S:\O\to[0,1]$ at $\bx\in\O$  is based only on the intensities in a neighborhood of $\bx$. Although the resulting denoised spectrogram  greatly improves the IF estimation of the signal, both computational time and low energy harmonic removing were drawbacks which motivated different approaches, see \cite{dfgv2008}.

In this paper we show that nonlocal filters such as Neighborhood filters are computationally more efficient to deal with these images and give similar results. The main idea in Neighborhood filters, see \cite{buades2005} for a detailed description,  is to  take into account in the filtering process the global gray scale values to define neighboring pixels. Thus,   the denoised value at pixel $\bx$ is defined as an  average of values at pixels having a gray scale value close to $S(\bx)$.
The resulting nonlocal algorithm is written as 
\begin{equation}
\label{def.NF}
 NF_h S(\bx)=\frac{1}{C(\bx)}\int_\O \exp\left(-\frac{\abs{S(\bx)-S(\by)}^2) }{h^2}\right) S(\by)d\by,
\end{equation}
where $C(\bx)=\int_\O \exp\left(-\abs{S(\bx)-S(\by)}^2) h^{-2}\right) d\by$ is a normalization factor. 

 An interesting variation of the Neighborhood filter algorithm is the 
 Yaroslavsky filter algorithm \cite{yarolavsky1985}, in which the domain of integration in \fer{def.NF} is restricted to a spatial neighborhood of the base pixel $\bx$, i.e. to a ball $B_{\rho}(\bx)=\left\{\by\in\O:d(\by, \bx)<\rho\right\}$, where $d$ is usually taken as the Euclidean distance. In this way, both local and nonlocal filtering are taken into account. Similar to the Yaroslavsky filter, SUSAN \cite{smith1997} and Bilateral \cite{tomasi1998} filters incorporate the 
 domain restriction as a weighted distance to the reference pixel
 \begin{equation}
 \label{def.Y}
  SNF_{h,\rho} S(\bx)=\frac{1}{C(\bx)}\int_\O \exp\left( -\frac{\abs{\by - \bx}^2}{\rho^2} - \frac{\abs{S(\bx)-S(\by)}^2) }{h^2}\right) S(\by)d\by.
 \end{equation}
 
 According to Buades et al. \cite{buades2006}, an interpretation of these filters may be given in terms of their asymptotic equivalence to differential diffusion filters,  based on the relative size of $h$ and $\rho$.
 When $h$ is much larger than $\rho$, $S$ is filtered by a heat equation. If $h$ and $\rho$ have the same order, the filter behaves as a filtering-enhancing algorithm depending on the magnitude of $\abs{\grad S}$, i.e., as a 
 nonlinear diffusion Perona-Malik type filter. In the last case, if $\rho$ is much larger than $h$, the signal is hardly modified.
 
 The denoising process is usually performed as an iteration of formula \fer{def.NF} by the following scheme 
\begin{equation}
\label{def.NFi}
  S_{i+1}(\bx)=\frac{1}{C_0(\bx)}\int_\O \exp\left(-\frac{\abs{S_0(\bx)-S_0(\by)}^2) }{h^2}\right) S_i(\by)d\by,
\end{equation}
with $C_0(\bx)=\int_\O \exp\left(-\abs{S_0(\bx)-S_0(\by)}^2) h^{-2}\right) d\by$.
In Singer et al. \cite{singer2009}, a probabilistic approach is used for the understanding of iterated Neighborhood and Nonlocal Means filters, see \fer{def.NL}-\fer{def.F} below.
Their rate of convergence towards the constant steady state and their blurring properties are analyzed interpreting the successive iterations \fer{def.NFi} as a random walk. The iterated composition of the corresponding transition probability matrix is then shown to produce the noise reduction at a certain asymptotic rate, acting specially on the high frequency components.

We finish this section by  describing an important property which is related to the special character of the spectrogram interpretation as an image. 
The Nonlocal Means algorithm uses, for a noisy image $v:\O\to\R$, the following regularization 
\begin{equation}
\label{def.NL}
 NL(v)(\bx)=\frac{1}{C(\bx)}\int_\O \exp\left(-\frac{F(\bx,\by) }{h^2}\right) v(\by)d\by,
\end{equation}
where $C(\bx)=\int_\O \exp\left(-F(\bx,\by) h^{-2}\right) d\by$ is the normalization factor, 
\begin{equation}
\label{def.F}
F(\bx,\by)=\int_\O G_\sigma(\bz)\abs{v(\bx+\bz)-v(\by+\bz)}^2d\bz,
\end{equation}
 and $G_\sigma$ is a Gaussian kernel with standard deviation $\sigma$.  The role of function $F$ is to weight the point-wise Euclidean distance based similarity of $v$ on  \emph{Gaussian neighborhoods} $B_\sigma(\bx)$ and $B_\sigma(\by)$ in the averaging procedure of \fer{def.NL}.
Relaxing this effect to perform, instead of point-wise, averaged Euclidean distance comparisons on Gaussian neighborhoods for $\WV(f;\cdot)$,  i.e. redefining $F$ as 
\begin{equation*}
F(\bx,\by)=\left| \int_\O G_\sigma(\bz) \left(\WV(f;\bx+\bz)-\WV(f;\by+\bz)\right) d\bz \right|^2,
\end{equation*}
we obtain $F(\bx,\by)=\abs{S(f;\bx)-S(f;\by)}^2$ , where $S(f;\cdot)$
is obtained by using the Gaussian $G_\sigma =\WV( \vfi_\sigma ; \cdot )$ 
as convolution kernel in \fer{specwv}, for a Gaussian window $\vfi_\sigma$ of zero mean and standard deviation $\sigma$, see \cite{mallat}. Therefore, considering the  regularization of $\WV(f;\cdot)$ via \fer{def.NL}-\fer{def.F}, we obtain  
\begin{equation}
\label{def.NLS}
 NL (\WV(f;\cdot))(\bx)=\frac{1}{C(\bx)}\int_\O \exp\left(-\frac{\abs{S(f;\bx)-S(f;\by)}^2,) }{h^2}\right) \WV (f;\by)d\by,
\end{equation}
which is in close relation to \fer{def.NFi} once the first iteration approximates the integrand $\WV(f;\cdot)$
to $S(f;\by)$. Thus, in a loose sense, the Nonlocal Means algorithm applied to the Wigner-Ville distribution of a signal is related to the direct application of the Neighborhood filter to the spectrogram of the signal, as obtained through a Gaussian window.


\section{Discretization and numerical experiments}

In this section we present numerical demonstrations on spectrograms of synthetic and biomedical signals of the nonlocal
Neighborhood filter \fer{def.NFi}, the Yaroslavsky filter \fer{def.Y} iterated as in \fer{def.NFi}, 
and a Total Variation based PDE local filter, see \fer{pde}-\fer{id} below. In order to compare the results, we use
an objective measure, the Mean Square Error (MSE) between the denoised and the clean signals, and a subjective measure based on the visual inspection of the spectrogram itself and on some related magnitudes: the IF lines and the time-energy content.

\subsection{Discretization}

Computation of the spectrogram is a standard operation
performed by applying the discrete fast Fourier transform (dfft)
to the convolution of the signal with the window, which we choose as a
zero mean unit variance Gaussian window. 
Once the spectrogram is produced,
it is normalized in some range $[0,Q]$, with $Q>0$, obtaining in this way 
a discrete function $S:\O\to[0,Q]$. The choice $Q=255$ corresponds to
the usual digital image range.

Let us consider the regular mesh of $\O=(0,T)\times(0,F)$, as produced by the spectrogram computation, given by $\left\{(t_m,\om_n)\right\}$, with $t_m=m T/(M+1)$, for $m=0,1,\ldots,M+1$ and $\om_n = nF/(N+1)$, for $n=0,1,\ldots,N+1$, and consider the corresponding barycenters 
$\bx_{m,n}$ for $m=0,1,\ldots,M$ and $n=0,1,\ldots,N$. To each $\bx_{m,n}$ we associate a surface element $dA=TF/MN$. 

For $Q\in \mathbb{N}$, we consider the $Q-$quantized spectrogram $S_0(\bx)=k$ if $S(\bx)\in[k-\frac{1}{2},k+\frac{1}{2})\cap [0,Q]$, for $k=0,1,\ldots,Q$,  inducing the partition of $\O$ given by
\begin{equation}
\label{partition}
 \O =  \bigcup_{k=0}^{Q} B_k,\qtext{with }B_k=\left\{\bx\in\O : S_0(\bx)=k\right\}.
\end{equation}
According to Eq. \fer{def.NFi} for the iteration of the Neighborhood filter, and using \fer{partition}, we define for $j=0,1,\ldots,Q$, and for the iteration index $i=0,1,\ldots$
\begin{align}
\label{f.1}
 S_{i+1}^{NF}(\left\{\bx\in B_j\right\})=& \frac{1}{C_0(\bx)}\int_\O \exp\left(-\frac{\abs{S_0(\bx)-S_0(\by)}^2) }{h^2}\right) S_i^{NF}(\by)d\by\\
 =& \frac{\sum_{k=0}^Q e^{-(\frac{j-k}{h})^2} \displaystyle\int_{B_k}  S_i^{NF}(\by)d\by}{\sum_{k=0}^Q e^{-(\frac{j-k}{h})^2} \abs{B_k}}, \nonumber
\end{align}
with $S_{0}^{NF}=S_0$. We consider the zero-order approximations
\begin{equation}
\label{int.app}
 \int_{B_k}  S_i^{NF}(\by)d\by \approx \sum_{\by_{m,n} \in B_k} S_i^{NF}(\by_{m,n}) dA
 ,\qtext{and } \abs{B_k} \approx \mathcal{C}(k) dA,
\end{equation}
where $\mathcal{C}(k)$ denotes the number on nodes contained in $B_k$. Relabeling $S_{i+1}^{NF}$ in terms of its level sets, the approximation to formula \fer{f.1} (for which we keep the same notation) reads  
\begin{align*}
 S_{i+1}^{NF}(j)= \frac{\sum_{k=0}^Q e^{-(\frac{j-k}{h})^2} \sum_{\by_{m,n} \in B_k} S_i^{NF}(\by_{m,n}) }{\sum_{k=0}^Q e^{-(\frac{j-k}{h})^2} \mathcal{C}(k)},
\end{align*}
or, in a more compact form
\begin{align*}
 S_{i+1}^{NF}(j)= \frac{< a(j,\cdot),\mathcal{S}^i >}{<a(j,\cdot), \mathcal{C}>},
\end{align*}
with
\begin{equation*}
 a(j,k)=e^{-(\frac{j-k}{h})^2},\quad \mathcal{S}^i(k)= \sum_{\by_{m,n} \in B_k} S_i^{NF}(\by_{m,n}),
\end{equation*}
and $<\cdot,\cdot>$ denoting the scalar product in $\R^{Q+1}$.

In a similar way, we may find the following expression for the iteration of the Yaroslavsky type filter given by \fer{def.Y}. However, for this filter we can not operate on terms of level sets, and a particular expression for each $\bx_{m,n}\in B_j$ must be given,
explaining the large time execution differences between Neighborhood and Yaroslavsky filtering:
\begin{align*}
 S_{i+1}^{Y}(\bx_{m,n})= \frac{\sum_{k=0}^Q e^{-(\frac{j-k}{h})^2} \displaystyle\int_{B_k}  e^{-(\frac{\by-\bx_{m,n}}{\rho})^2}S_i^{Y}(\by)d\by}{\sum_{k=0}^Q e^{-(\frac{j-k}{h})^2} \displaystyle\int_{B_k}  e^{-(\frac{\by-\bx_{m,n}}{\rho})^2}d\by}. \nonumber
\end{align*}
Approximating the integral terms as in \fer{int.app}, we obtain
\begin{align}
\label{f.2}
 S_{i+1}^{Y}(\bx_{m,n})= \frac{\sum_{k=0}^Q e^{-(\frac{j-k}{h})^2} \sum_{\by_{p,q}\in B_k}  e^{-(\frac{\by_{p,q}-\bx_{m,n}}{\rho})^2}S_i^{Y}(\by_{p,q})}{\sum_{k=0}^Q e^{-(\frac{j-k}{h})^2} \sum_{\by_{p,q}\in B_k}  e^{-(\frac{\by_{p,q}-\bx_{m,n}}{\rho})^2}}. \nonumber
\end{align}

Finally, for comparison porpouses, we used the following denoising algorithm based on a
convection-diffusion nonlinear PDE related to that introduced by Rudin et al. \cite{rudin1992}.
The problem is written as:
find $S:[0,\frac{1}{\eps}]\times \O\to [0,Q]$ such that
\begin{eqnarray}
&& \frac{\partial S}{\partial \tau}+\frac{\eps}{2}
\grad\log(S_{0*})\cdot\grad S -  \Div \Big(\frac{\grad S}{\abs{\grad
S}}\Big)=0 \qtext{in } [0,\frac{1}{\eps}]\times\O  \label{pde} ,
 \\[0.1cm]
&&\grad S \cdot \mathbf{n}=0 \qtext{on }   
[0,\frac{1}{\eps}]\times\partial\O, \label{bc} \\
&&S(0,\cdot)=S_0 \qtext{in } \O , \label{id}
\end{eqnarray}
where $S_{0*}=G_\sigma*S_0$, with $G_\sigma$ a Gaussian regularizing kernel of variance $\sigma$.
Here, $\eps>0$ controls the strength of transport towards the maxima of $S_0$,
in relation to the  mean curvature diffusion term. The addition of the transport
term in \fer{pde}, as well as the bounded \emph{time} domain
$[0,\frac{1}{\eps}]$, which are not usual in image denoising algorithms, are
motivated by the notion of differential reassignment introduced by Auger et al.
\cite{auger1997}, which is proper of  one-dimensional signals, see
\cite{dfgv2007b} for further details. 

The discretization of \fer{pde}-\fer{id}
follows the standard Finite Element methodology for the total variation
denoising problem, see \cite{dfgv2008b}. We consider a time semi-implicit Euler scheme and a $\mathbb{P}_1$
continuous finite element approximation in space.  In particular,
for the approximation, $S_{k+1}$, at time $t=t_{k+1}$, we
use the diffusive coefficients obtained in the previous time step
$t_k$, that is,
\begin{equation}\label{8}
\frac{S_{k+1}-S_k }{\delta t}+\frac{\epsilon}{2} \nabla
\log(S_{0*}) \cdot \nabla S_k -\Div
\Big(\frac{1}{\sqrt{|\nabla S_k|^2+\tilde\epsilon^2}}\nabla S_{k+1}\Big)=0,
\end{equation}
for some $\tilde\epsilon >0$. The convective term is computed by an upwind scheme, evaluating
for each node the derivatives in the element corresponding to the
upwind direction. Therefore, for each time step, we solve an
elliptic problem with a diffusive term coefficient bounded from
above.

As mentioned in the introduction to this section, we use two measures to compare the denoising 
results:
\begin{enumerate}
 \item \emph{The Mean Square Error} between the clean spectrogram, $S_c(\bx)$, and the filtered spectrogram after $I$ iterations, $S_I(\bx)$
 \begin{equation}
  \text{MSE}=\frac{\nor{S_c-S_I}}{\nor{S_c}},
 \end{equation}
where $\nor{\cdot}$ denotes the $L^2$ norm of a matrix.

\item \emph{The visual inspection of the IF lines.}  We use a simple algorithm to produce
candidates to IF lines of the corresponding spectrograms. 
We consider the truncation $v(t,\om)=S(t,\om)$ if  $S(t,\om)\geq
\beta$, and $v(t,\om)=0$ elsewhere, with $\beta =\mean _\O {(S)}$
in the experiments. For each $t\in[0,T]$ we find the $N$
connected components of the set $\left\{\om\in (0,F):
v(t,\om)>0\right\}$, say $C_n (t)$, for $n=1,\ldots,N(t)$, and
define the function
\begin{equation}
\label{def.IF}
\IF(t,n)=\frac{\displaystyle\int_{C_n (t)} \om v(t,\om)d\om}{\displaystyle\int_{C_n (t)}
v(t,\om)d\om},
\end{equation}
which is the frequency gravity center of the component $C_n (t)$.
In this way, we shrink each connected component to one point to
which we assign the average image intensity through the function
$\INT(t,n)=\mean _{C_n (t)}{(v(t,\cdot))}$. Finally, we plot
function $\IF$ only for components with averaged intensity,
$\INT$, greater than a certain threshold, $i\in [0,Q]$. This
final image does not seem to be very sensible under small
perturbations of the parameters $\beta$ and $i$.

\end{enumerate}

\begin{table}[t]
\centering
\begin{tabular}{|c|c|c|c|c|}
\hline 
\multicolumn{2}{|c|}{} & Neighborhood & Yaroslavsky & Nonlinear diff.\tabularnewline
\hline 
\hline 
\multirow{2}{*}{Experiment 1} & MSE & 0.28 & 0.16 & 0.17\tabularnewline
\cline{2-5} 
 & Ex. time & 2.33 & 874 & 255\tabularnewline
\hline 
Experiment 2  & Ex. time & 0.24 & 95 & 22\tabularnewline
\hline 
Experiment 3  & Ex. time & 1.03 & 125 & 32\tabularnewline
\hline 
\end{tabular}
\label{table}
\caption{Mean square error (MSE) and execution time (seconds), of the different algorithms and 
experiments. A smaller MSE indicates that the estimate is closer to the original image. The numbers
have to be compared on each row.} 
\end{table}


In all the experiments we have chosen the usual image quantization into $256$ levels ($Q=255$).
For the Neighborhood and Yaroslavsky filters, the variance of the Gaussian neighborhood for the level 
lines is taken as $h=10$. For the latter, we fix the variance of the local neighborhood as $\rho=10$ so,
according to \cite{buades2005}, the nonlinear diffusion regime applies.
For the Neighborhood algorithm, the iteration stopping criterium is established in terms
of the relative difference between two consecutive iterations, i.e. we stop the algorithm if
\begin{equation*}
 \frac{\nor {S_{i+1}-S_i}_{L^2}}{\nor {S_i}_{L^2}} < {\rm tol},
\end{equation*}
with ${\rm tol}$ empirically chosen to perform few iterations. For the PDE based filter algorithm we stop when
the artificial time $\tau_{end}=1/\eps$ is reached. The parameter $\eps$, controlling the 
ratio transport-diffusion is empirically chosen. For the Yaroslavsky filter we use the same number of iterations
as for the PDE algorithm. In Table~\ref{table}, the execution time
expressed in seconds refers to the {\tt tic-toc} Matlab command for a script run under a 
Core i7 (quad) processor, built on a standard laptop.

\no\textbf{Experiment 1. } We use a one second $4$KHz synthetic
signal composed by the addition of two signals. The first is the
addition of pure tones and chirps:
\[
x_1(t)=c_1\big(\sin (2\pi 500t) +\sin (2\pi 700t)+\sin (2\pi 1000t^2)+
\sin (2\pi 600t^3)\big),
\]
while the second, $x_2$, is a uniformly distributed real random
variable. We normalize them to have $\nor{x_i}_{L^2}=1$ (so the
constant $c_1$) and define the test signal as $x=x_1+x_2$, i.e.,
with $\SNR=0$. We then compute the spectrogram, $S$, for a Gaussian window of unit variance.
The size of the corresponding image is $2048\times 512$ pixels. Other parameters are chosen as 
follows: ${\rm tol}=0.04$, $\eps=0.02$ and $d\tau=2.5$,  implying
$\tau_{end}=50$.

\begin{figure}
\centering
\includegraphics[width=6cm,height=4cm]{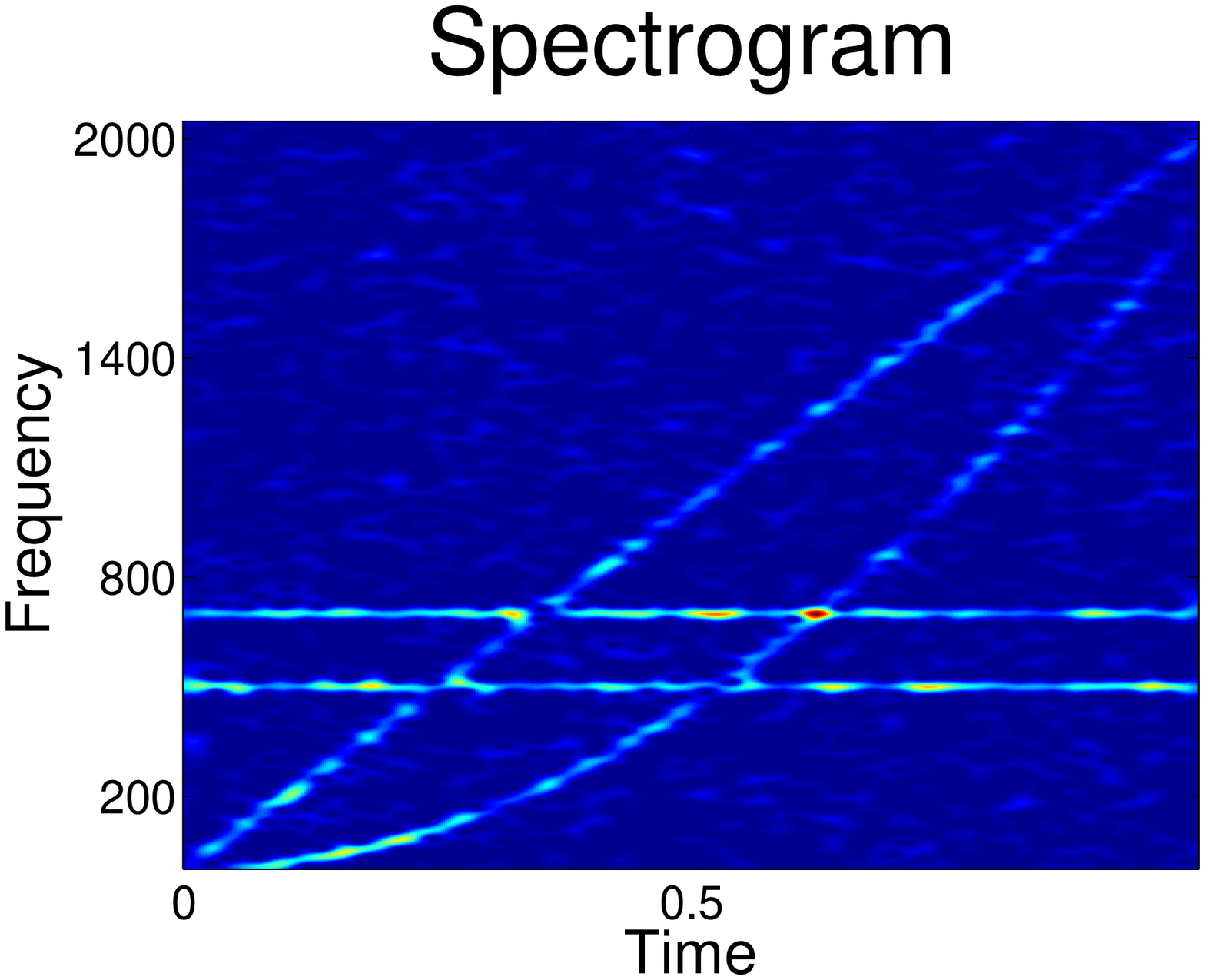}
\includegraphics[width=6cm,height=4cm]{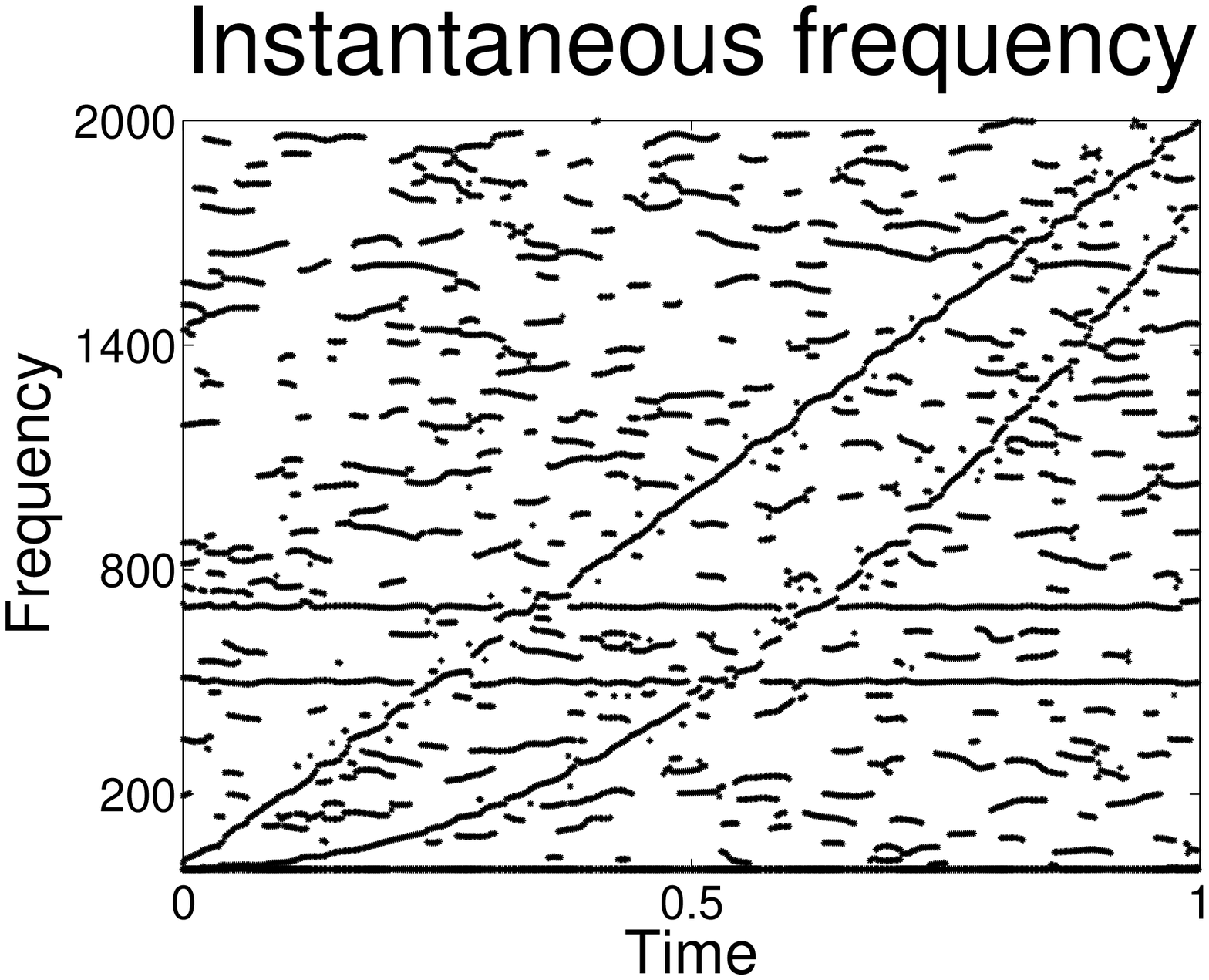}\\
\includegraphics[width=6cm,height=4cm]{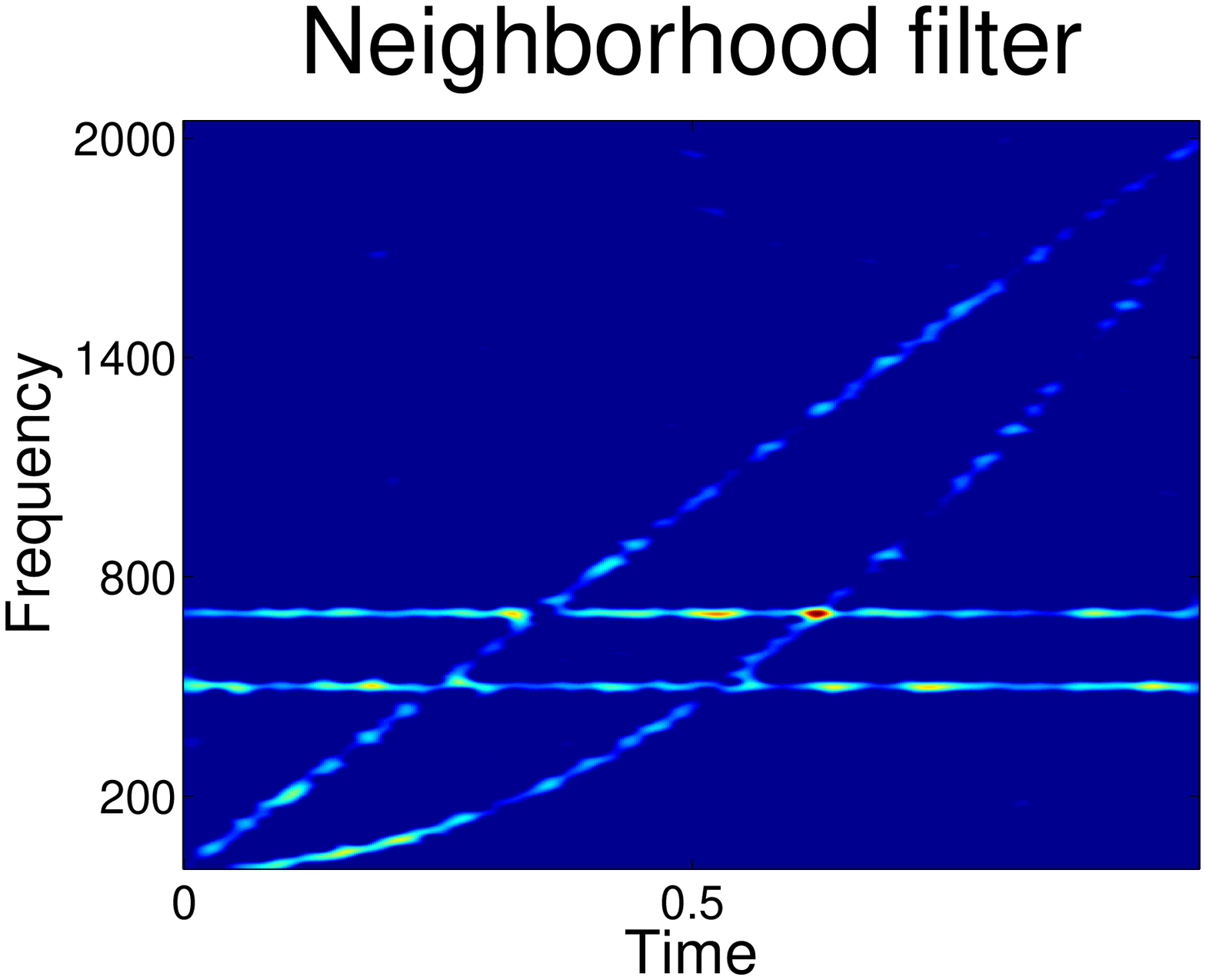}
\includegraphics[width=6cm,height=4cm]{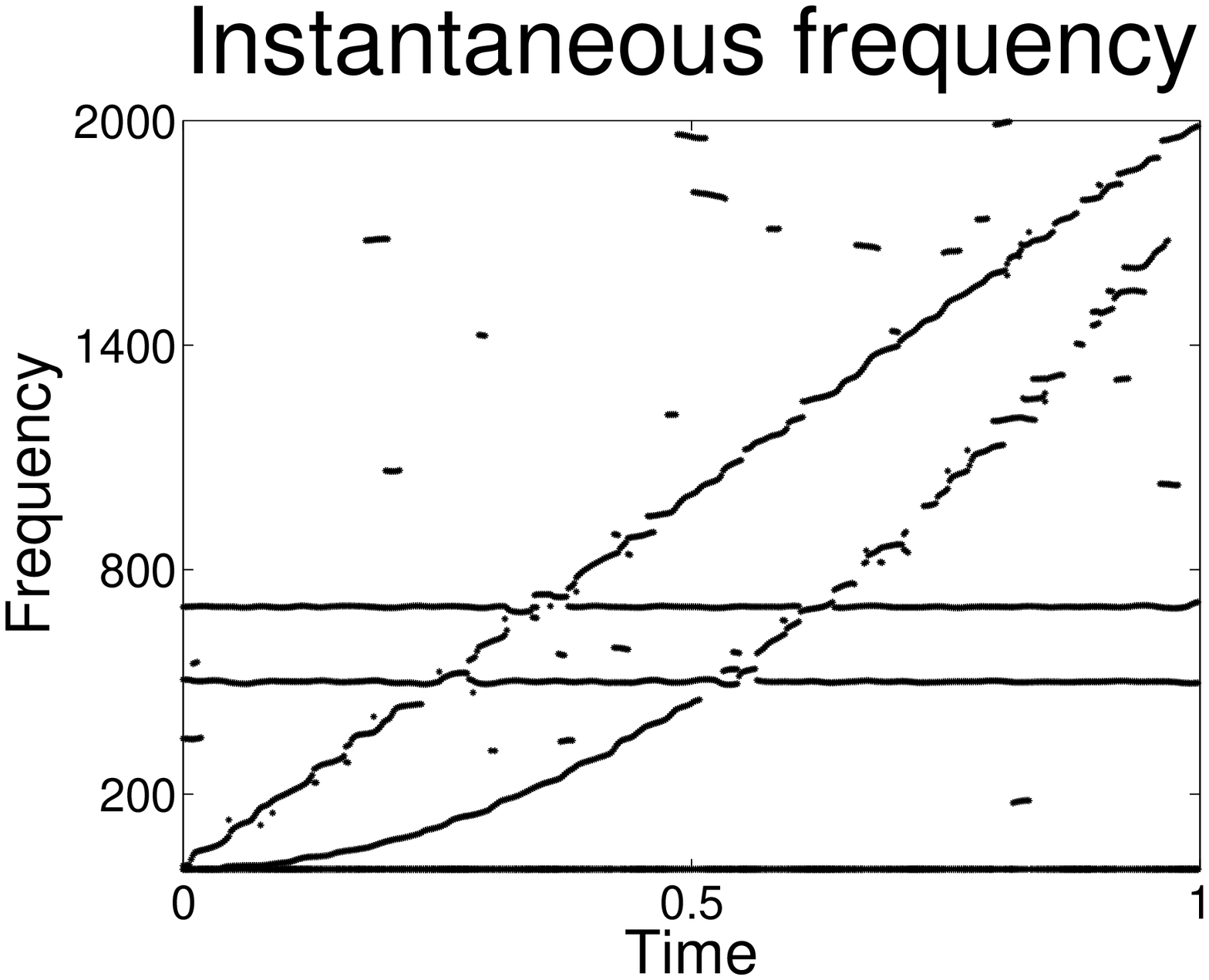}\\
\includegraphics[width=6cm,height=4cm]{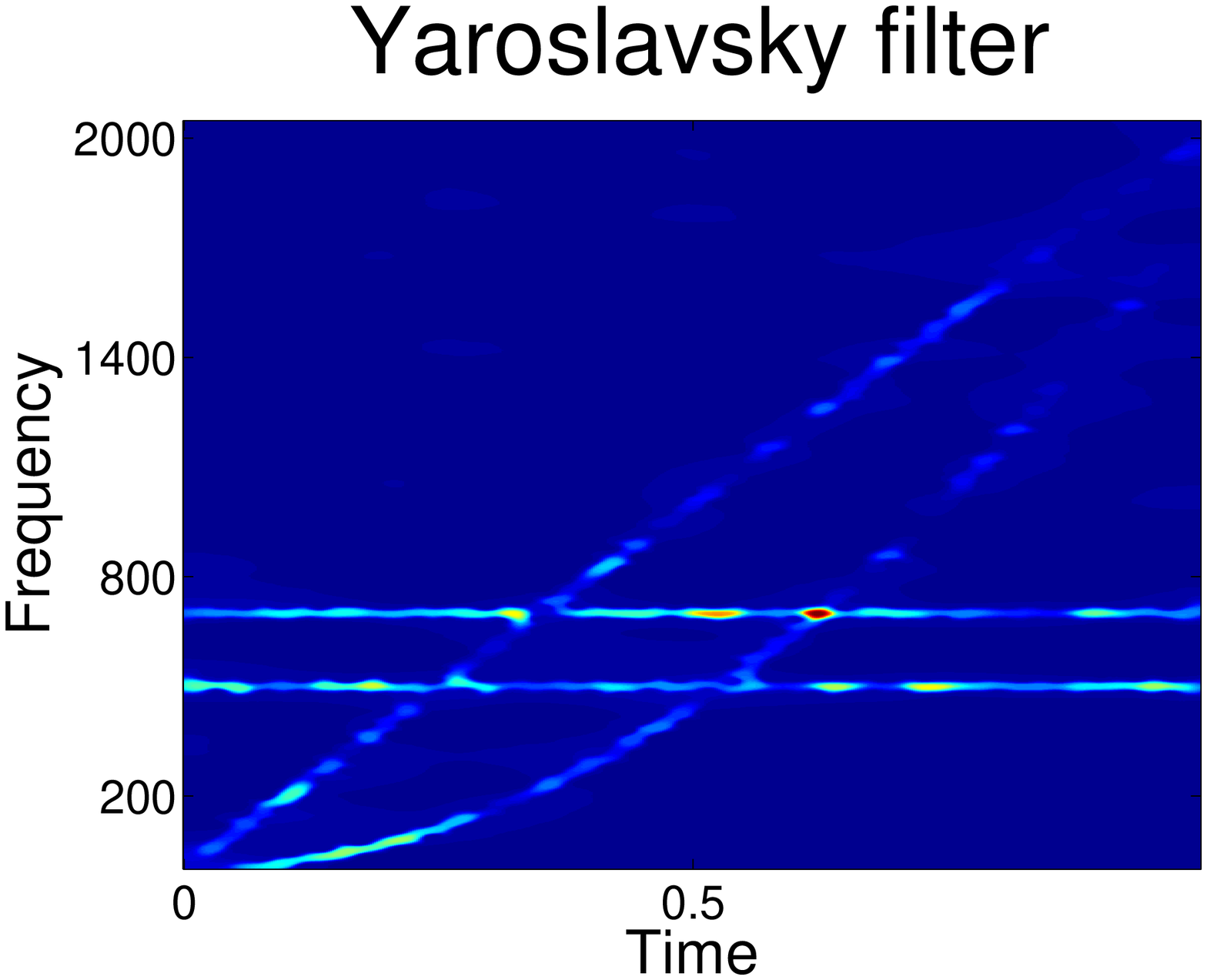}
\includegraphics[width=6cm,height=4cm]{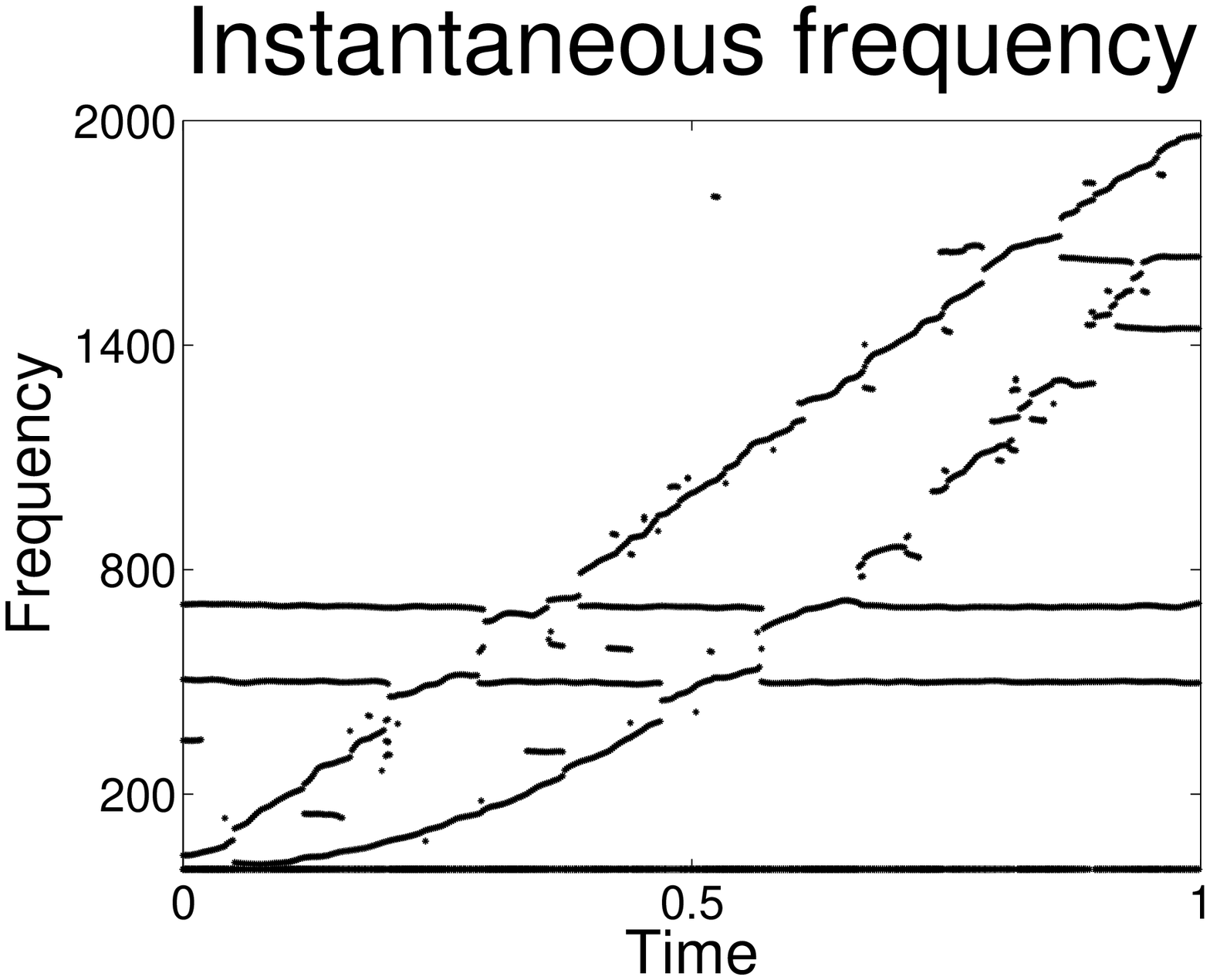}\\
\includegraphics[width=6cm,height=4cm]{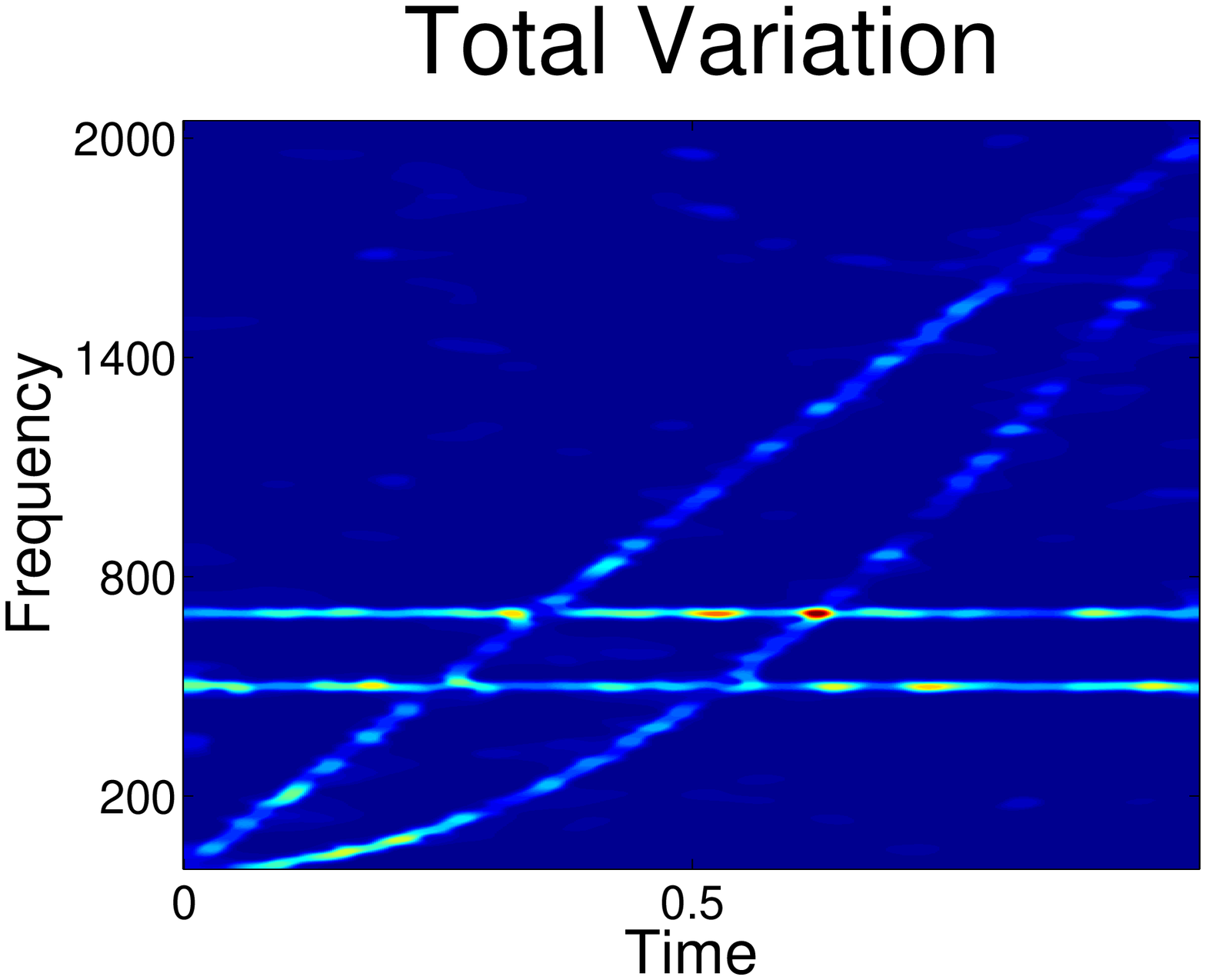}
\includegraphics[width=6cm,height=4cm]{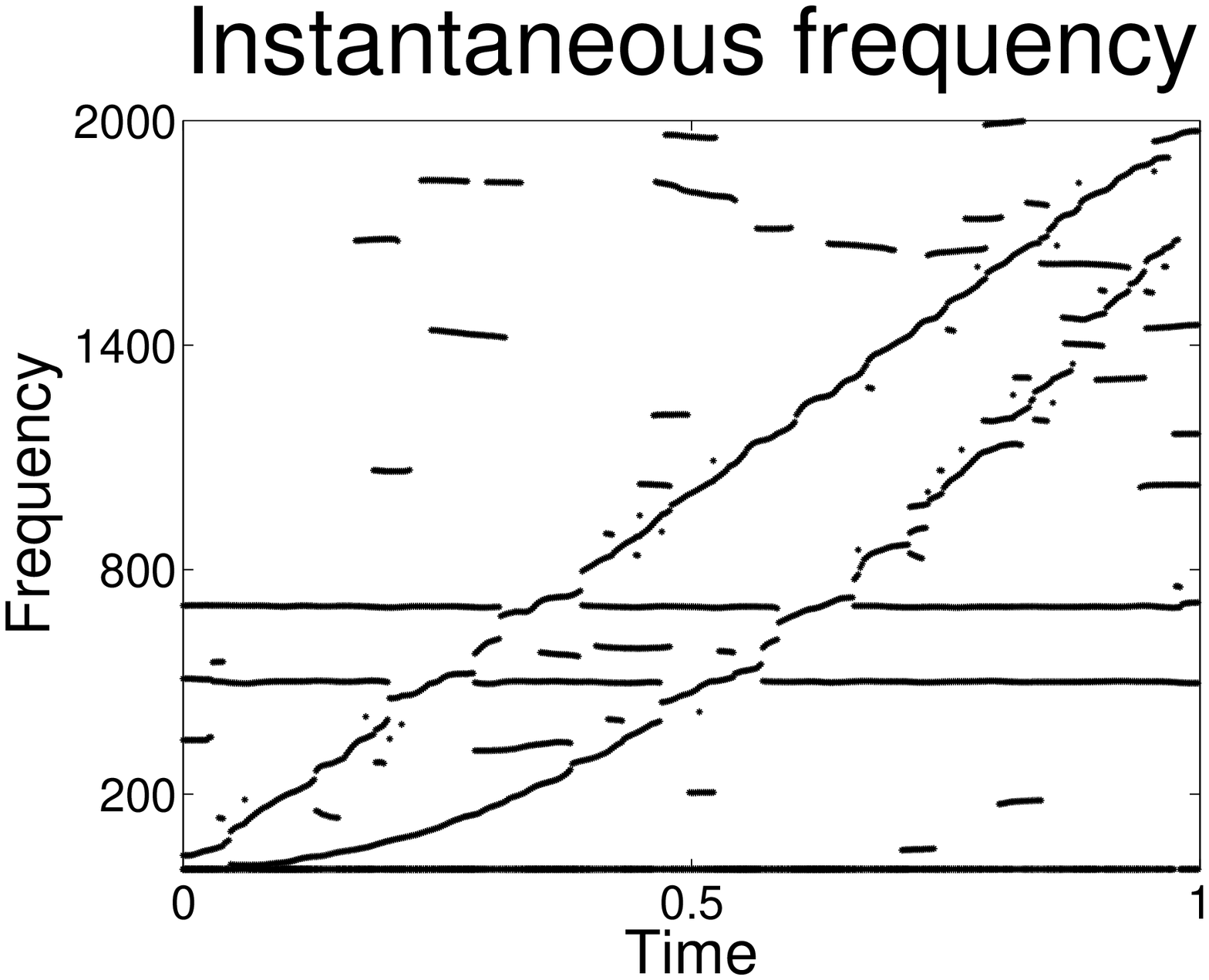}\\
\caption{\label{fig1}{\small Experiment 1. Left: spectrogram of the noisy signal and 
filtered images obtained with the different algorithms. Right: corresponding IF lines according
to formula \fer{def.IF}.}}
\end{figure}

In Figure~\ref{fig1} we show the graphic results. The left panel contains the spectrogram of 
the noisy signal and the corresponding images filtered with the three described algorithms.
In comparison with the original spectrogram, all filtered images are very closed from each other.
We observe that the image obtained by Neighborhood filtering keeps more noise than the others, but
also experiments less diffusive effects. This is specially noticeable in the broken IF segments 
appearing at high frequencies, where the Yaroslavsky and the TV filtering diffuse locally while
the Neighborhood filter just remove the information of this area. The similarity of these denoised
images is also checked by comparing their corresponding MSE, see  Table~\ref{table}. However, 
the most impressive result is that concerning the execution times which gives a ratio larger 
than 200 in favor of the Neighborhood algorithm.

\bigskip

\no\textbf{Experiment 2. }We use a recording done in wilderness \cite{luis1},  
from where we extracted a signal of $2.25$
seconds which is affected by a strong background noise. Since
field data recorders are set to 44.1 KHz meanwhile wolves signals
are rarely out of the range $200-3000$ Hz, we start by filtering
and down-sampling the signal to speed up computations. The resulting spectrogram 
image size is $172\times 690$ pixels. Other parameters are chosen as 
follows: ${\rm tol}=0.01$, $\eps=0.2$ and $d\tau=0.25$,  implying
$\tau_{end}=5$.

\begin{figure}
\centering
\includegraphics[width=6cm,height=4cm]{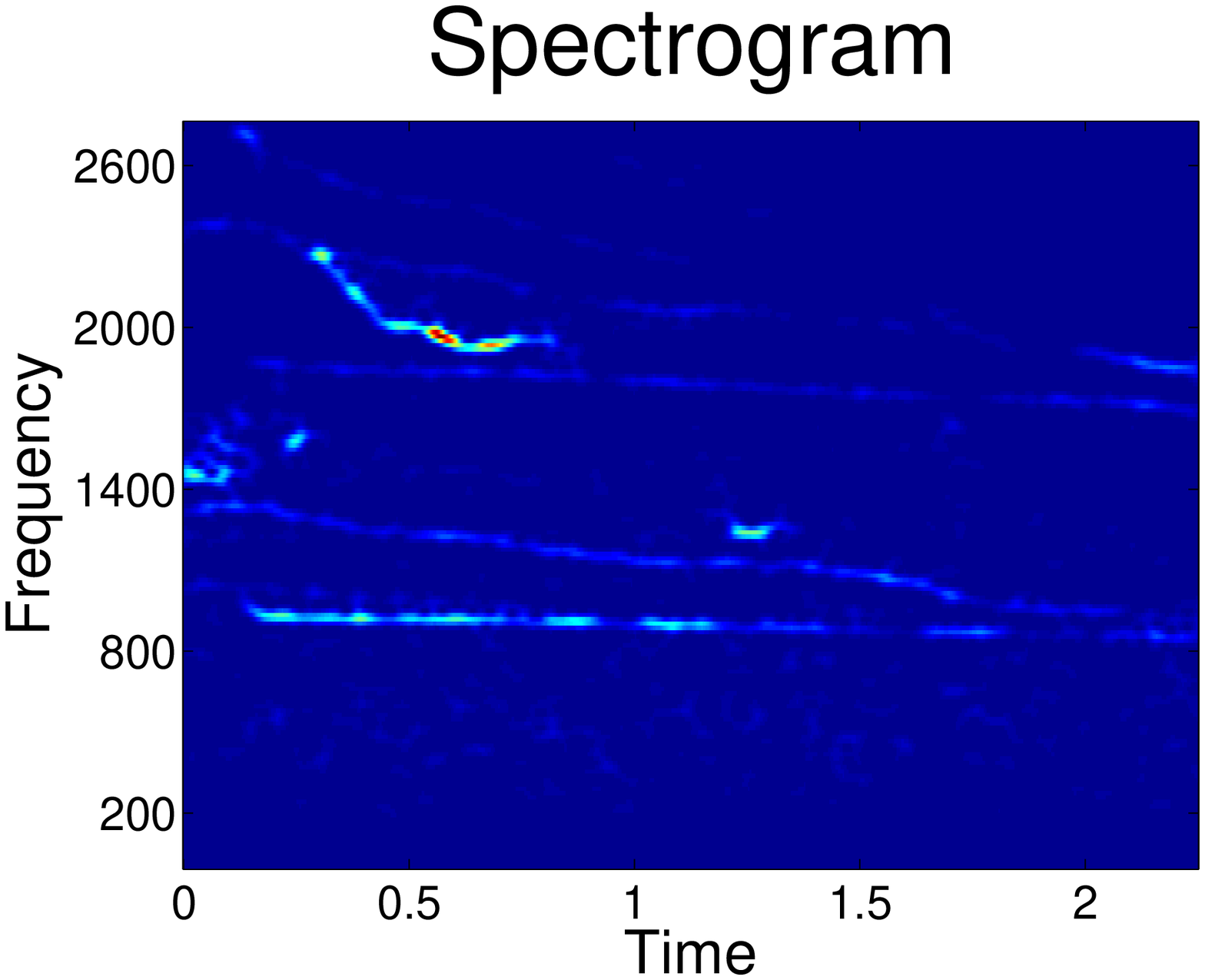}
\includegraphics[width=6cm,height=4cm]{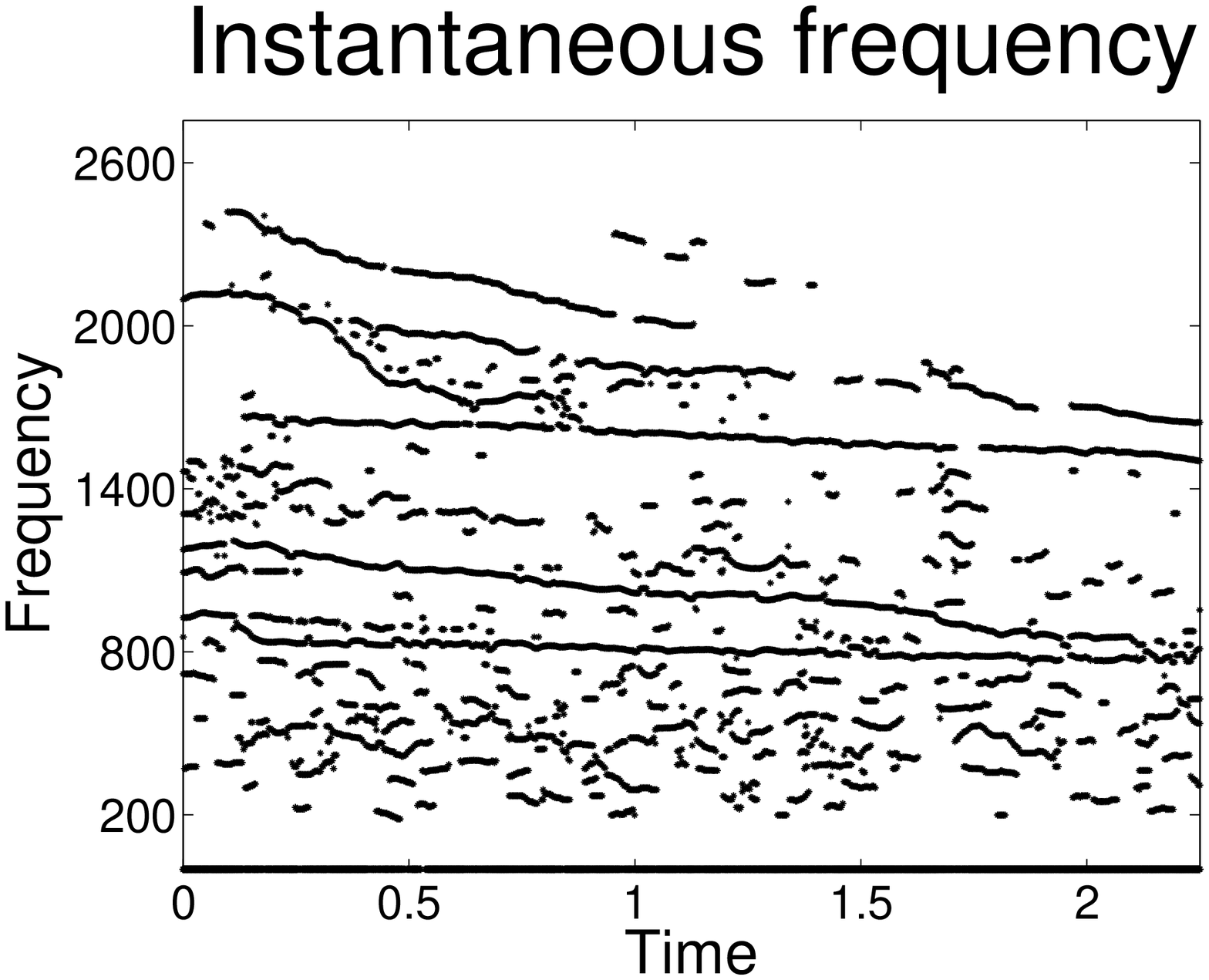}\\
\includegraphics[width=6cm,height=4cm]{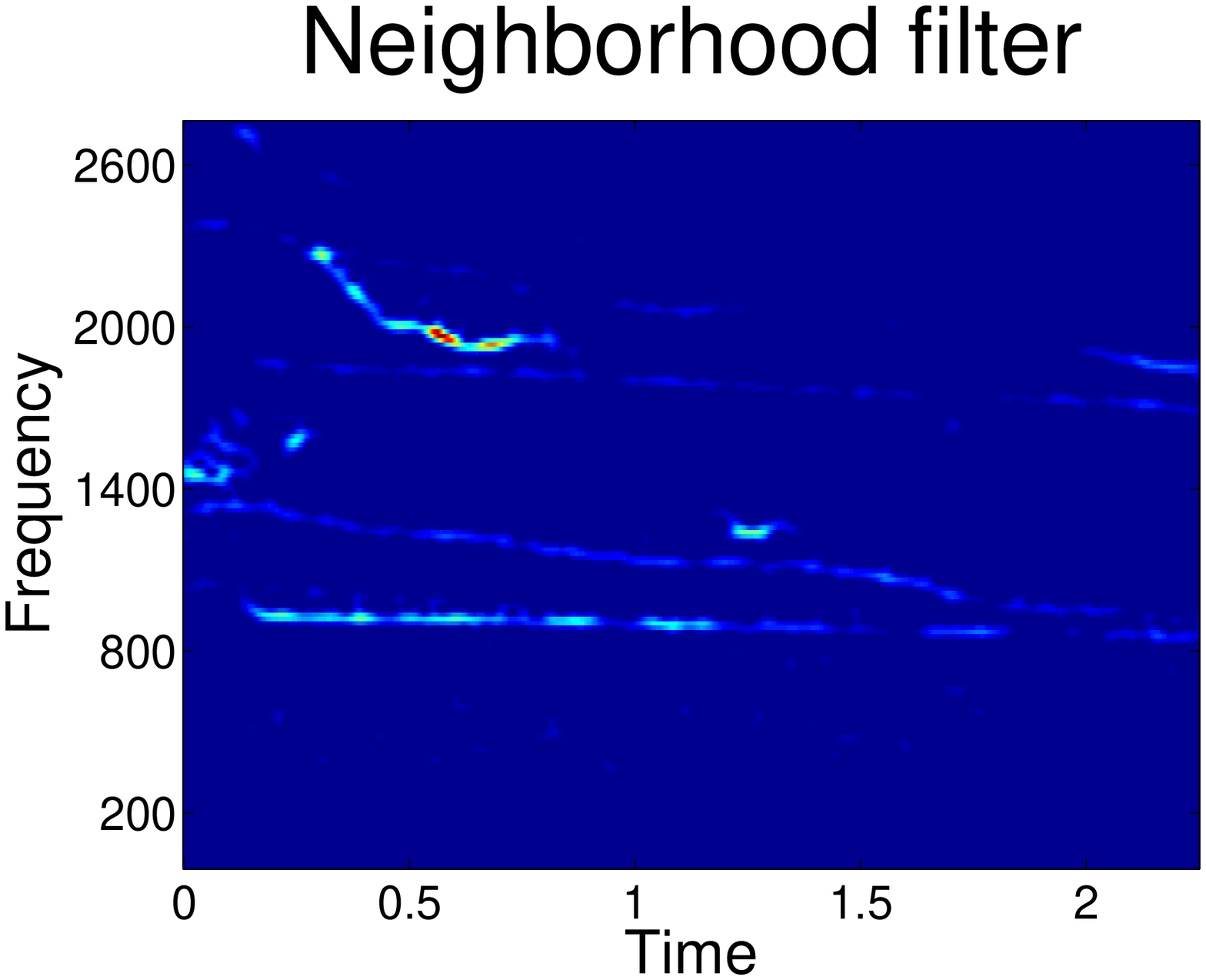}
\includegraphics[width=6cm,height=4cm]{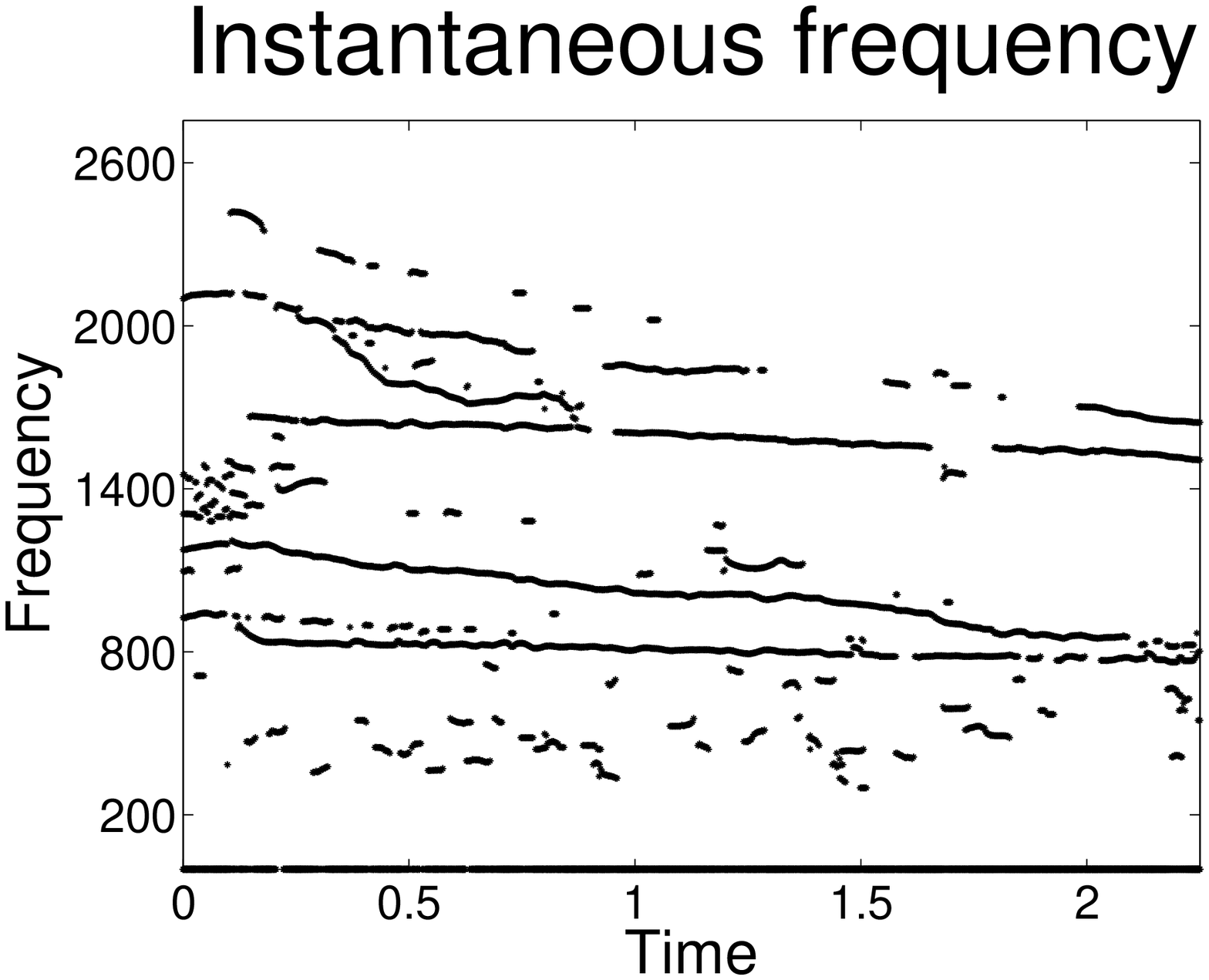}\\
 \includegraphics[width=6cm,height=4cm]{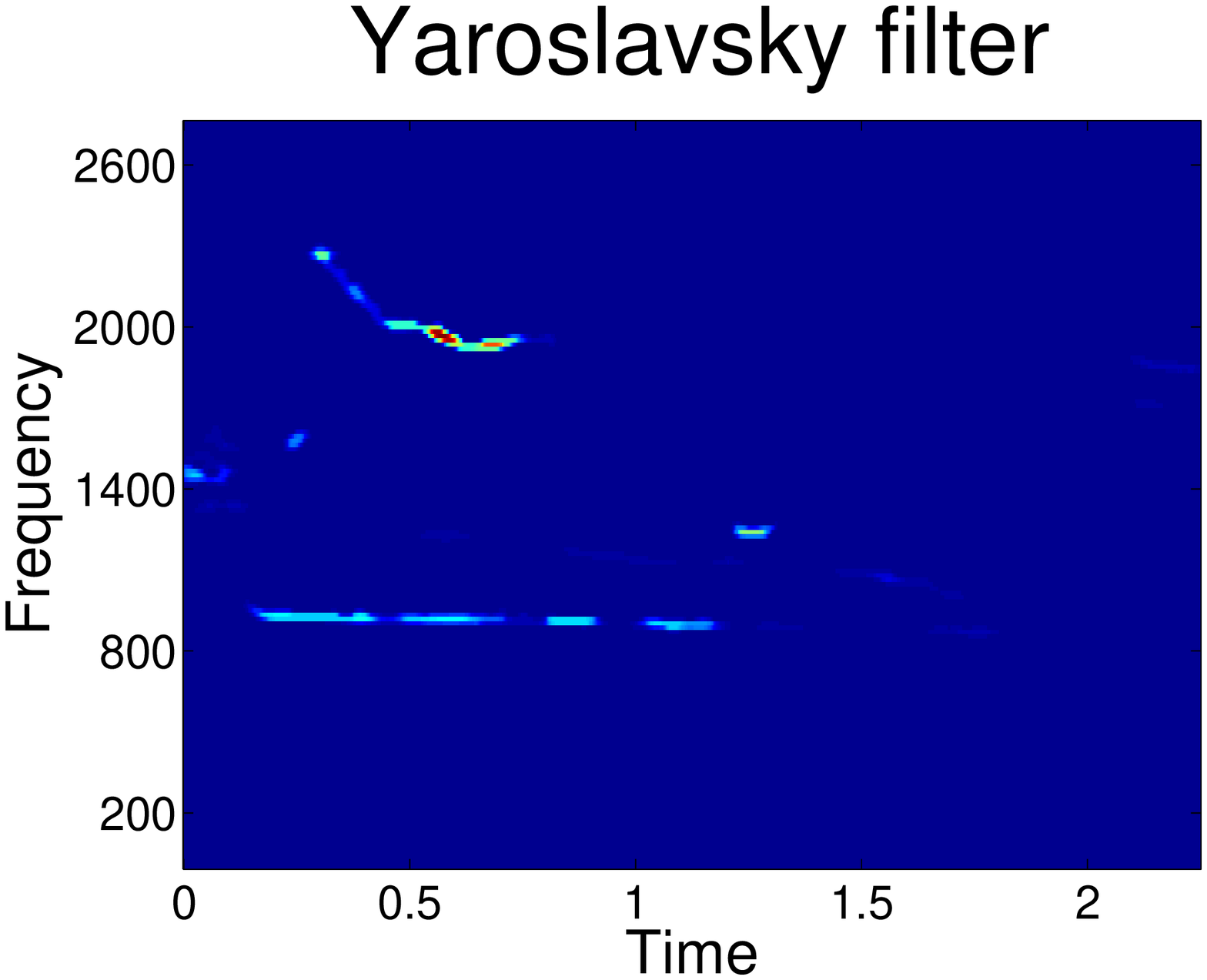}
 \includegraphics[width=6cm,height=4cm]{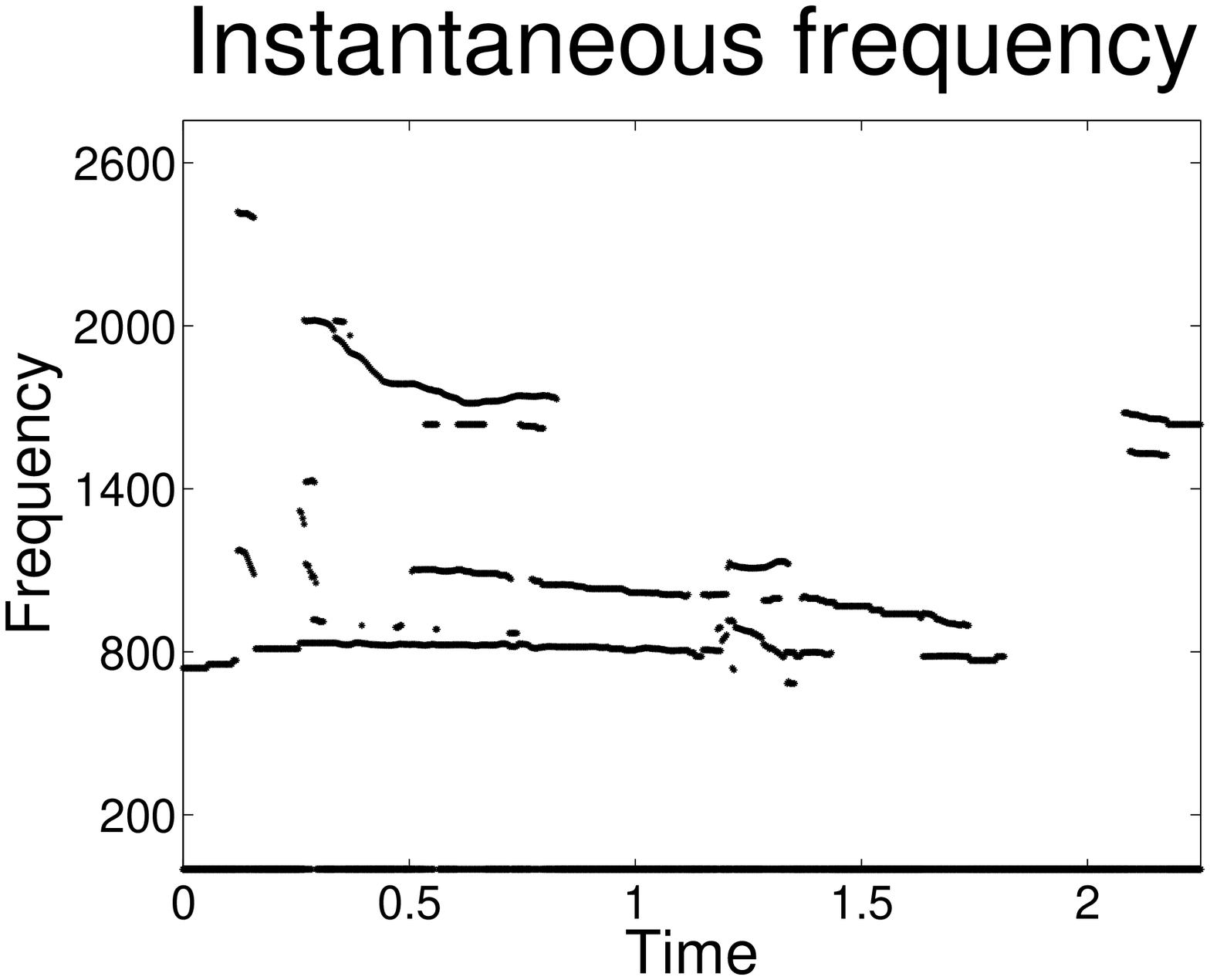}\\
 \includegraphics[width=6cm,height=4cm]{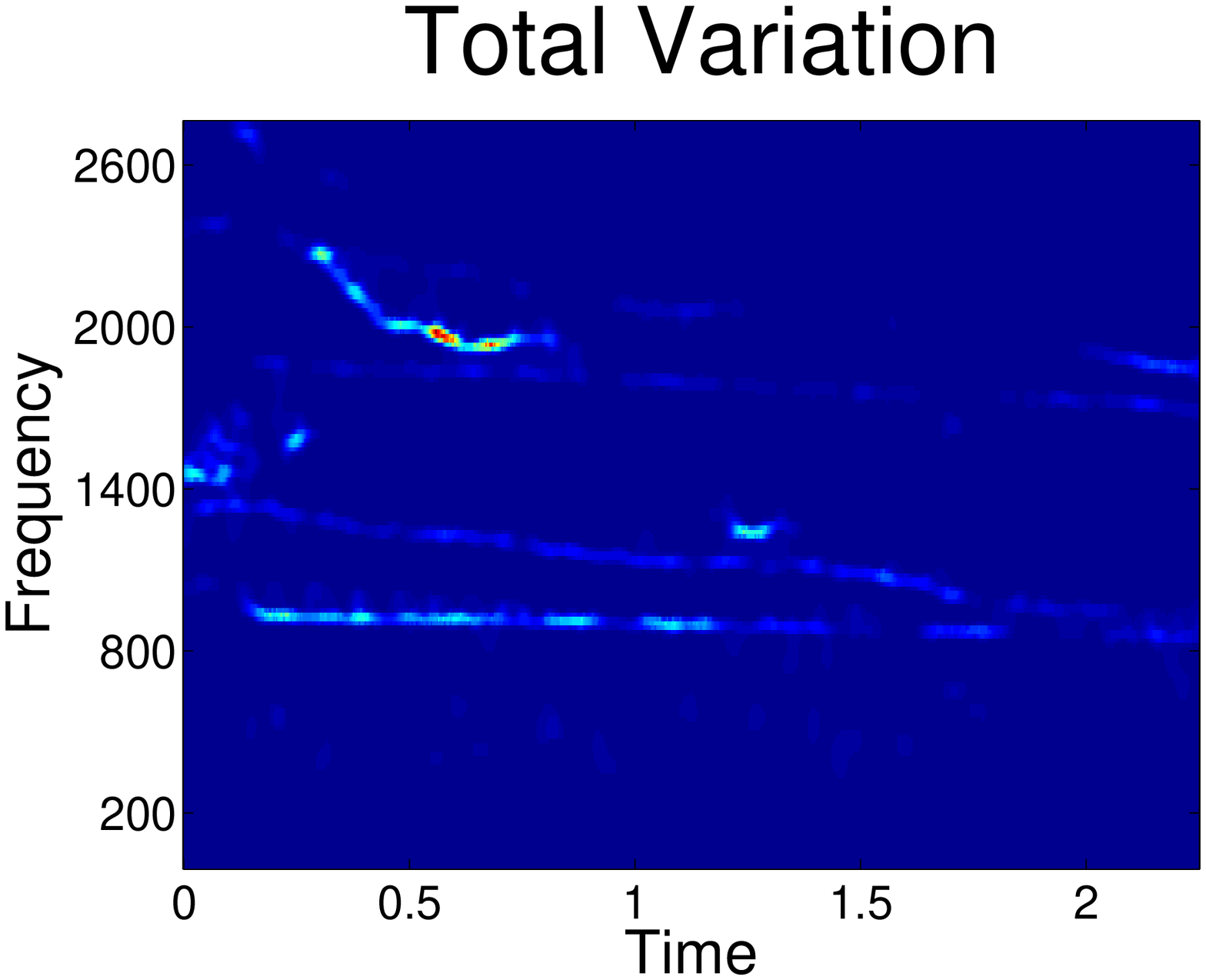}
 \includegraphics[width=6cm,height=4cm]{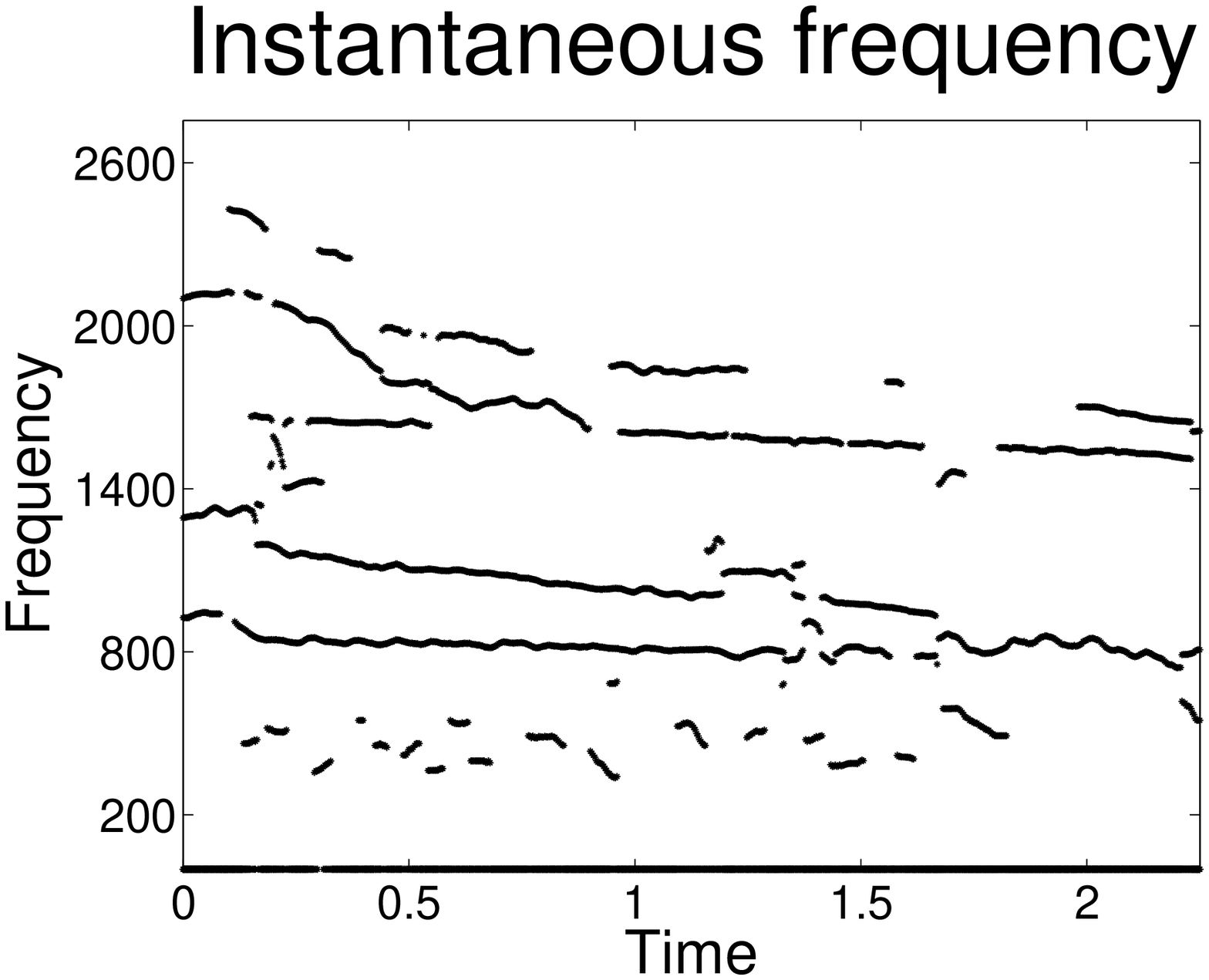}
 \caption{\label{fig2}{\small Experiment 2. Left: spectrogram of the noisy signal and 
filtered images obtained with the different algorithms. Right: corresponding IF lines according
to formula \fer{def.IF}.  }}
\end{figure}

In Figure~\ref{fig2} we show the graphic results. Again, the left panel contains the spectrogram of 
the signal and the results of filtering. In this case, the noise is not Gaussian and the filtering
algorithms are less adapted to the problem. The Neighborhood filter keeps most of the relevant information
(the fairly straight lines) but leaves part of the low frequency noise unfiltered. The Yaroslavsky filter
removes the noise, but also the IF lines with less energy content. In a middle way, the TV algorithm 
removes most of the noise and keeps a large part of the IF lines, being for this example the most adequate
algorithm. However, execution times show again large differences in favor of the Neighborhood filter.

\bigskip

\no\textbf{Experiment 3. }
We used ECG signals obtained from 
the MIT-BIH Noise Stress Test Database \cite{moody1984,goldberger2000}, from which
we extracted a segment of 240 seconds (minutes 4 to 8), containing an arrhythmia episode in its 
central part.

The recording was made using physically active volunteers and standard ECG recorders.
Noise was added beginning after the first 5 minutes of each record, during two-minute segments alternating with two-minute clean segments. The signal-to-noise ratio was 12dB. The added noise was of the type
\emph{electrode motion artifact}, which is generally considered the most troublesome, since it can mimic 
the appearance of ectopic beats and cannot be removed easily by simple filters, as can noise of other types.

\begin{figure}
\centering
\includegraphics[width=4.3cm,height=4cm]{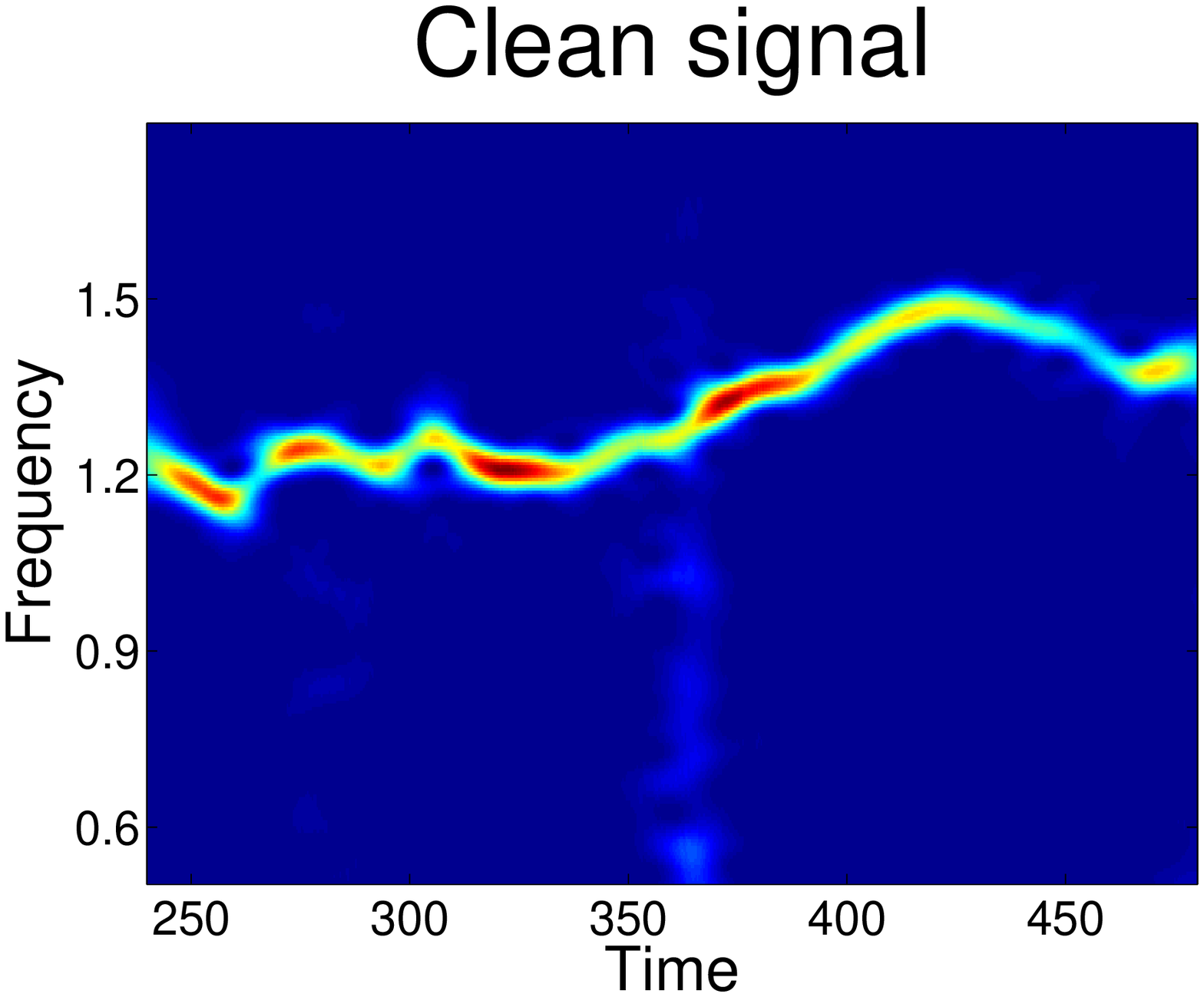}
\includegraphics[width=4.3cm,height=4cm]{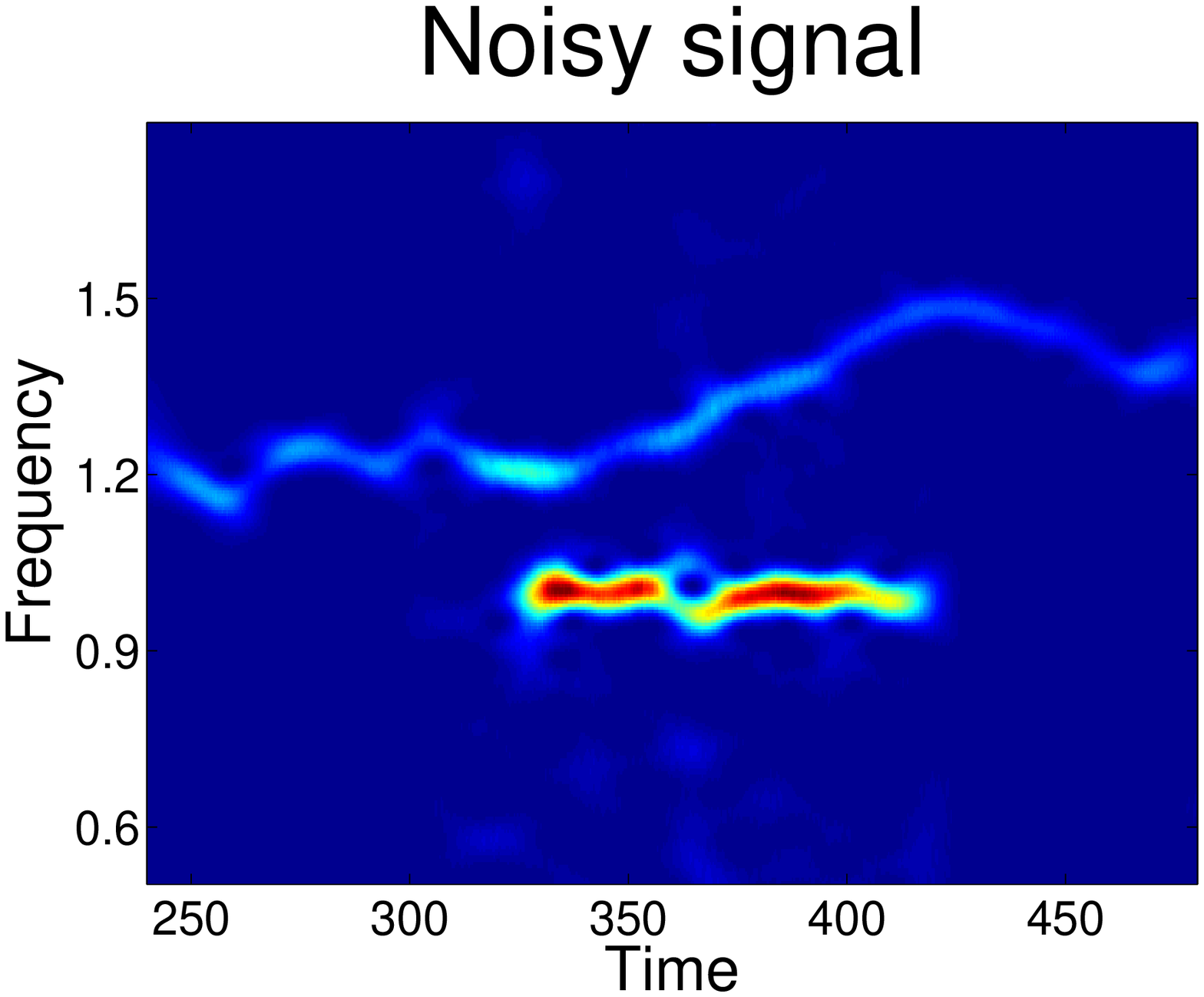}
\includegraphics[width=4.3cm,height=4cm]{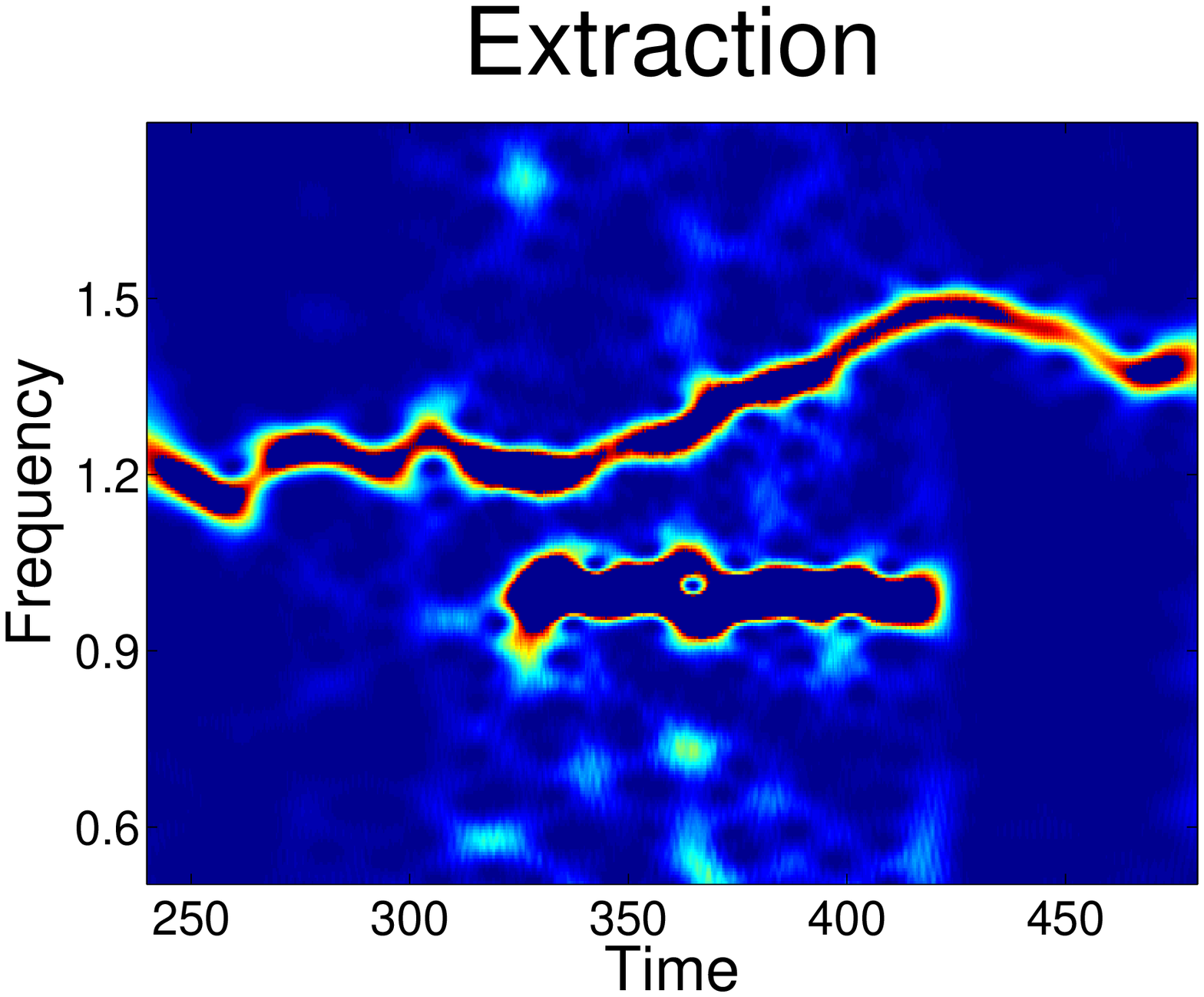}\\
\includegraphics[width=4.3cm,height=4cm]{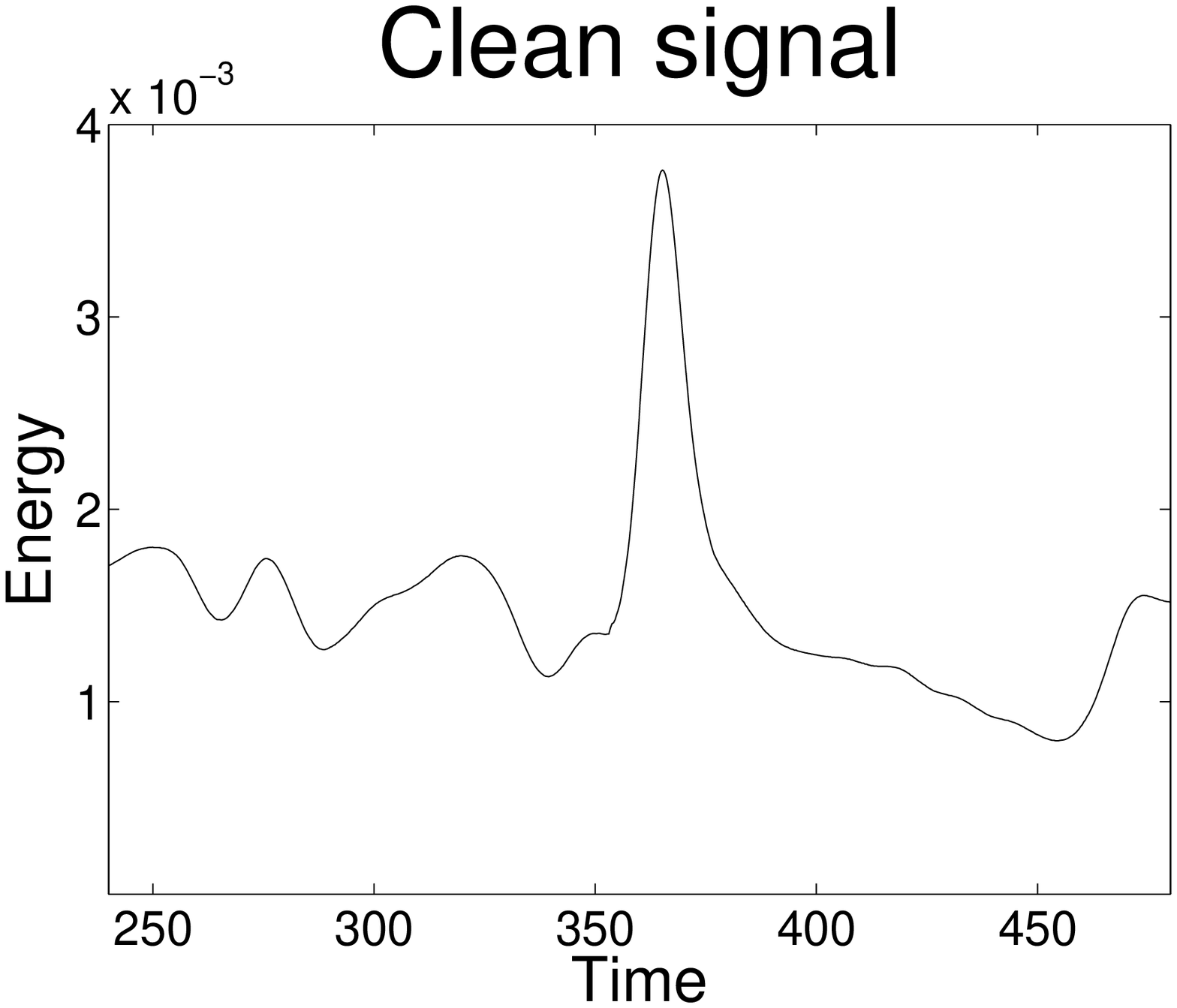}
\includegraphics[width=4.3cm,height=4cm]{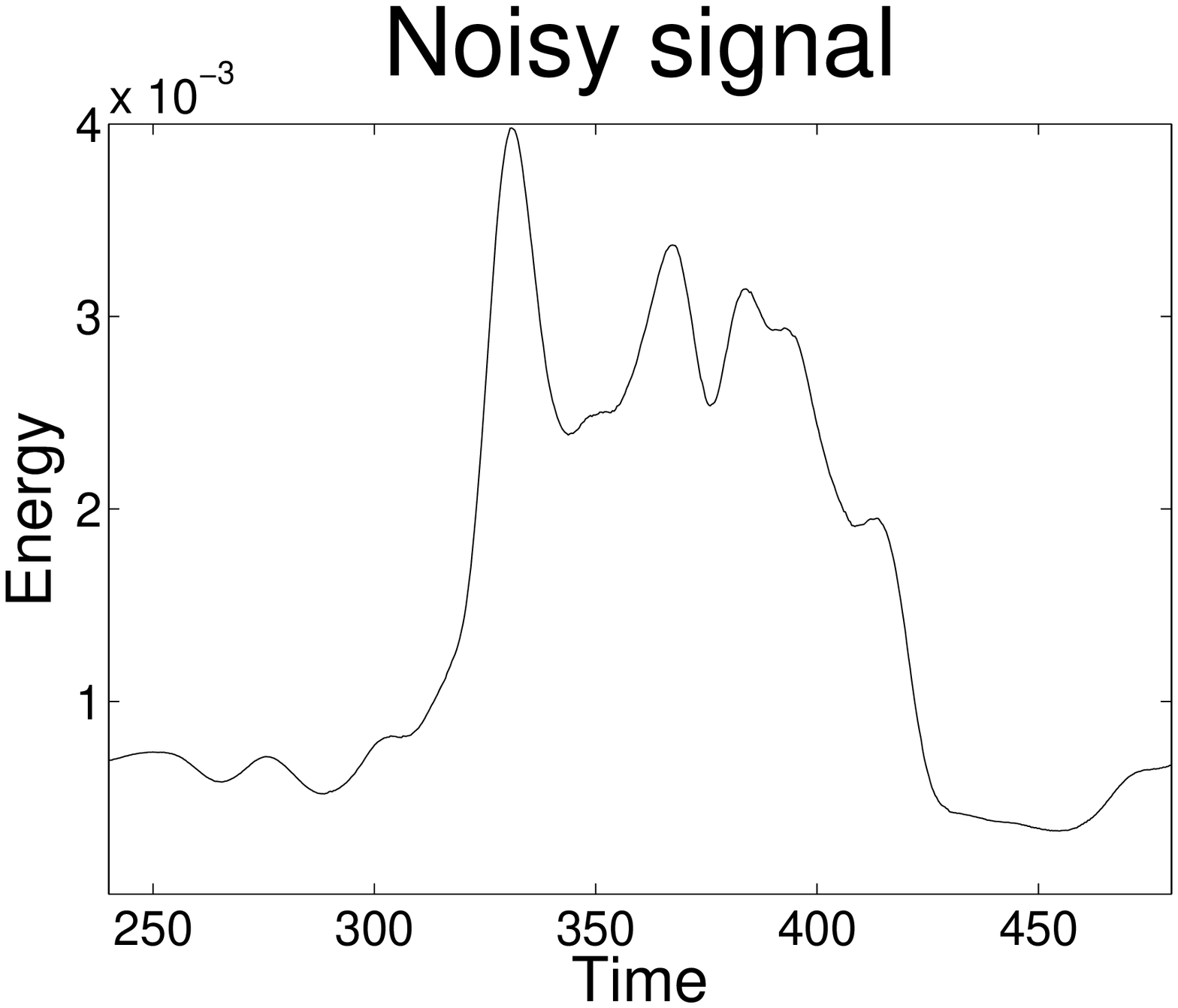}
\includegraphics[width=4.3cm,height=4cm]{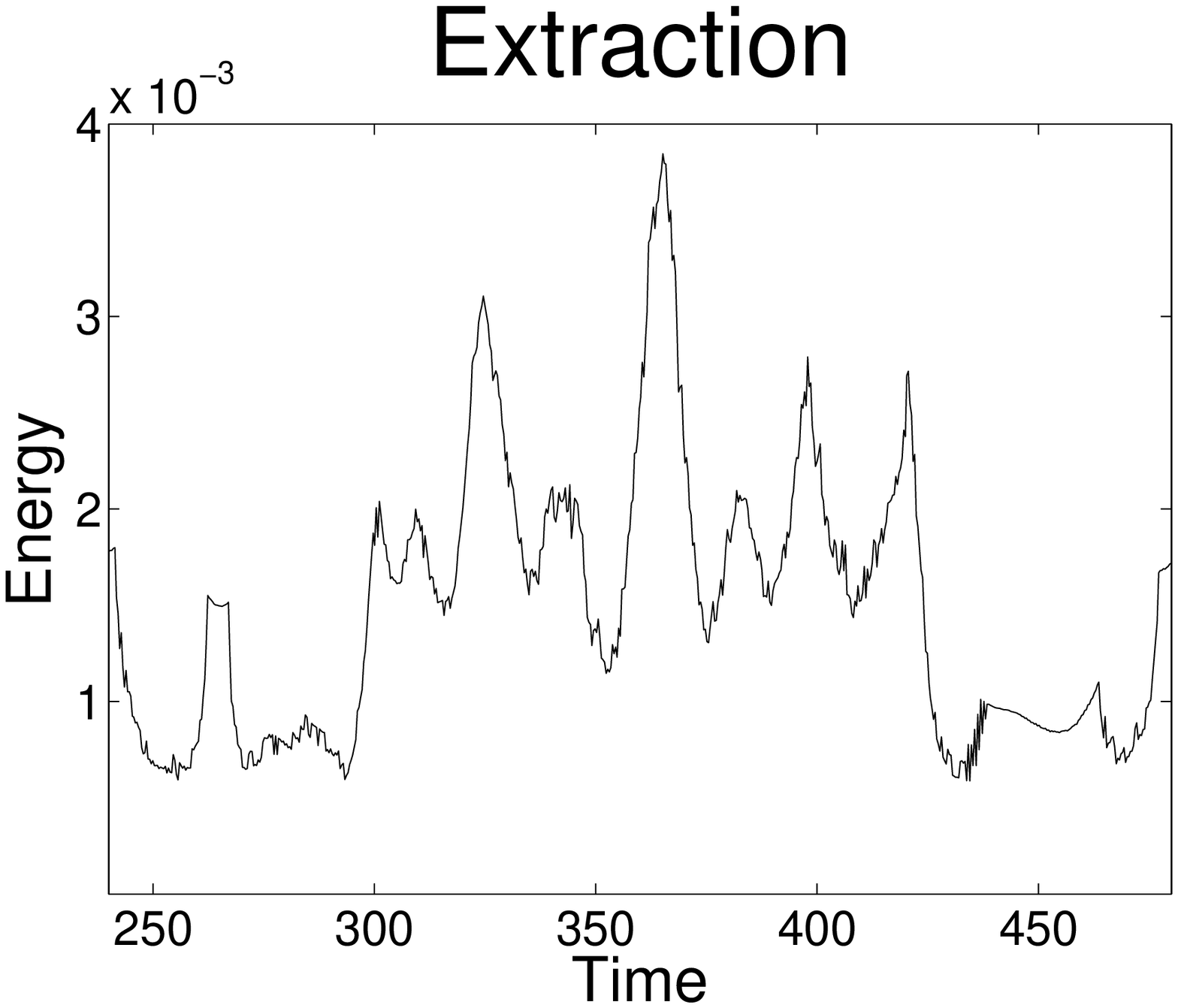}
\caption{\label{fig3}{\small Experiment 3. Recovery of an arrhythmia episode detection 
from a noisy signal.  }}
\end{figure}

For the clean image, see Fig. \ref{fig3} (first row), the spectrogram (size $236\times 676$) shows the usual energy spread of arrhythmia episodes around 
$t=360$. Integration of the energy in the frequency domain shows a pick at the same time instant, which allows 
to detect the episode, see Fig. \ref{fig3} (second row). In the central column of the same figure we 
repeat these calculations for the noisy signal. As it may be observed, the energy pick has been hidden due to the
time coincidence with the high energy noise. In the last column, we show the result of a filtering-extraction procedure.
Let $S_0$ be the spectrogram of the noisy signal and $S_n$ the image resulting from the application of the Neighborhood filter. Then we substract $S_n > \alpha$ from $S_0$, for some small $\alpha \in [0,Q]$ ($\alpha=1$ in the experiment). The corresponding image and 
its integration in the frequency domain are shown in the last column of Fig. \ref{fig3}. As we see, after the 
substraction of the most energetic components of the noisy signal, the noise and the normal beat frequency line, 
the arrhythmia episode may be again detected.
Although not shown, the results with the other algorithms are similar, being again in this experiment much larger
than that of the Neighborhood filter algorithm.






\begin{thebibliography}{}

 \bibitem{alm1992}
 L. \'Alvarez, P. L. Lions, J. M. Morel, Image selective smoothing
 and edge detection by nonlinear diffusion. II, SIAM J. Numer.
 Anal. 29(3) (1992) 845-866.



\bibitem{buades2005}
A. Buades, B. Coll, J. M. Morel, 
A review of image denoising algorithms, with a new one,
Multiscale Model. Simul. 4 (2005) 490--530.


\bibitem{buades2006}
A. Buades, B. Coll, J. M. Morel, 
Neighborhood filters and PDE's, 
Numer. Math. 105 (2006) 1--34.

\bibitem{buades2010}
A. Buades, B. Coll, J. M. Morel, 
Image Denoising Methods. A New Nonlocal Principle,
SIAM Rev. 52(1) (2010) 113--147.



\bibitem{auger1997}
E. Chassandre-Mottin, I. Daubechies, F. Auger, P. Flandrin,
Differential Reassignment, IEEE Signal Proc. Let. 4(10)
(1997) 293--294.



\bibitem{dfg2007}
B. Dugnol, C. Fern\'andez, G. Galiano, Wolves counting by
spectrogram image processing, Appl. Math. Comput. 186(1) (2007)
820--830.

\bibitem{dfgv2007b}
B. Dugnol, C. Fern\'andez, G. Galiano, J. Velasco,
Implementation of a diffusive differential reassignment
method for signal enhancement: An application to wolf population counting,
Appl. Math. Comput. 193 (2007) 374--384.

\bibitem{dfgv2008}
B. Dugnol, C. Fern\'andez, G. Galiano, J. Velasco,
On a chirplet transform based method applied to separating and 
counting wolf howls,
Signal Proc. 88 (2008) 1817--1826.

\bibitem{dfgv2008b}
B. Dugnol, C. Fern\'andez, G. Galiano, J. Velasco,
Evolution nonlinear diffusion-convection PDE models for spectrogram enhancement,
in Numerical Analysis and Applied Mathematics 1048 (2008) 166--169, 
AIP Conference Proceedings. 


\bibitem{goldberger2000}
A. L. Goldberger, L. A. N. Amaral, L. Glass, J. M. Hausdorff, P Ch. Ivanov, R. G. Mark, J. E. Mietus,
G. B. Moody, C. K. Peng, H. E. Stanley, PhysioBank, PhysioToolkit, and PhysioNet: 
Components of a New Research Resource for Complex Physiologic Signals. 
Circulation 101 (2000) e215--e220. Circulation Electronic Pages: \newline
{\tt http://circ.ahajournals.org/cgi/content/full/101/23/e215} (June 13). 

\bibitem{hopp1998} S.L. Hopp, M. J. Owren, C. S. Evans (Eds.),
Animal Acoustic Communication. Sound Analysis and Research Methods,
Springer, Berlin, 1998.


\bibitem{lee1983}
J.S. Lee,
Digital image smoothing and the sigma filter, 
Comput. Vision Graph. 24 (1983) 255--269.


\bibitem{luis1}
L. LLaneza, V. Palacios, Asesores en Recursos Naturales, S.L.,
Field recordings obtained in wilderness in Asturias (Spain) in the
2003 campaign.


\bibitem{mallat}
S. Mallat, A wavelet tour of signal processing, Academic Press,
London, 1998.

\bibitem{moody1984}
G. B. Moody, W.E. Muldrow, R. G. Mark,
A noise stress test for arrhythmia detectors,
Comput. Cardiol. 11 (1984) 381--384. 


\bibitem{rudin1992}
L. Rudin, S. Osher, E. Fatemi, 
Nonlinear total variation based noise removal algorithms,
Physica D 60 (1992) 259--268.


\bibitem{singer2009}
A. Singer, Y. Shkolnisky, B. Nadler,
Diffusion interpretation of Nonlocal Neighborhood Filters for signal denoising,
SIAM J. Imaging Sci. 2 (2009) 118--139. 


\bibitem{skonhoft}
A. Skonhoft, The costs and benefits of animal predation: An
analysis of Scandinavian wolf re-colonization, Ecol.
Econ. 58(4) (2006) 830--841.


\bibitem{smith1997}
S. M. Smith, J. M. Brady, 
SUSAN--a new approach to low level image processing, 
Internat. J. Computer Vision, 23 (1997) 45--78.

\bibitem{sornmo2005}
L. S\"ornmo, P. Laguna,
Bioelectrical Signal Processing in Cardiac and Neurological Applications,
Academic Press, 2005.


\bibitem{tomasi1998}
C. Tomasi, R. Manduchi, 
Bilateral filtering for gray and color images, 
in Proceedings of the 6th International Conference 
on Computer Vision (1998) 839--846.

\bibitem{xu2008}
H. Xu, Z.-H. Tan, P. Dalsgaard, B. Lindberg,
Robust speech recognition by Nonlocal Means denoising processing,
IEEE Signal Proc. Let. 15 (2008) 701--704.

\bibitem{yarolavsky1985}
L. P. Yaroslavsky, 
Digital Picture Processing. An Introduction, 
Springer-Verlag, Berlin, 1985.











\end{thebibliography}
\end{document}